\PassOptionsToPackage{unicode}{hyperref}
\PassOptionsToPackage{hyphens}{url}
\documentclass[
]{article}
\usepackage{amsmath,amssymb}
\usepackage{iftex}
\ifPDFTeX
  \usepackage[T1]{fontenc}
  \usepackage[utf8]{inputenc}
  \usepackage{textcomp} 
\else 
  \usepackage{unicode-math} 
  \defaultfontfeatures{Scale=MatchLowercase}
  \defaultfontfeatures[\rmfamily]{Ligatures=TeX,Scale=1}
\fi
\usepackage{lmodern}
\ifPDFTeX\else
\fi
\IfFileExists{upquote.sty}{\usepackage{upquote}}{}
\IfFileExists{microtype.sty}{
  \usepackage[]{microtype}
  \UseMicrotypeSet[protrusion]{basicmath} 
}{}
\makeatletter
\@ifundefined{KOMAClassName}{
  \IfFileExists{parskip.sty}{%
    \usepackage{parskip}
  }{
    \setlength{\parindent}{0pt}
    \setlength{\parskip}{6pt plus 2pt minus 1pt}}
}{
  \KOMAoptions{parskip=half}}
\makeatother
\usepackage{xcolor}
\usepackage[margin=1in]{geometry}
\usepackage{longtable,booktabs,array}
\usepackage{calc} 
\usepackage{etoolbox}
\makeatletter
\patchcmd\longtable{\par}{\if@noskipsec\mbox{}\fi\par}{}{}
\makeatother
\IfFileExists{footnotehyper.sty}{\usepackage{footnotehyper}}{\usepackage{footnote}}
\makesavenoteenv{longtable}
\usepackage{graphicx}
\makeatletter
\def\maxwidth{\ifdim\Gin@nat@width>\linewidth\linewidth\else\Gin@nat@width\fi}
\def\maxheight{\ifdim\Gin@nat@height>\textheight\textheight\else\Gin@nat@height\fi}
\makeatother
\setkeys{Gin}{width=\maxwidth,height=\maxheight,keepaspectratio}
\makeatletter
\def\fps@figure{htbp}
\makeatother
\usepackage{bbm}
\usepackage{amsmath}
\usepackage{amssymb}
\usepackage{graphicx}
\usepackage{bm}
\ifLuaTeX
  \usepackage{selnolig}  
\fi
\usepackage{bookmark}
\IfFileExists{xurl.sty}{\usepackage{xurl}}{} 
\hypersetup{
  pdftitle={Bridging the Unavoidable A Priori},
  pdfauthor={Peter S. Hovmand1,2; Kari O'Donnell3,1; Callie Ogland-Hand1,4; Brian Biroscak1; Douglas D. Gunzler3,4},
  hidelinks,
  pdfcreator={LaTeX via pandoc}}

\title{Bridging the Unavoidable A Priori\thanks{Prepared for the 43rd Conference of the International System Dynamics Society in Boston, United States.}}
\usepackage{etoolbox}
\makeatletter
\providecommand{\subtitle}[1]{
  \apptocmd{\@title}{\par {\large #1 \par}}{}{}
}
\makeatother
\subtitle{A Framework for Comparative Causal Modeling}
\author{Peter S. Hovmand\textsuperscript{1,2} \and Kari O'Donnell\textsuperscript{3,1} \and Callie Ogland-Hand\textsuperscript{1,4} \and Brian Biroscak\textsuperscript{1} \and Douglas D. Gunzler\textsuperscript{3,4}}
\date{July 31, 2025}

\begin{document}
\maketitle
\begin{abstract}
AI/ML models have rapidly gained prominence as innovations for solving previously unsolved problems and their unintended consequences from amplifying human biases. Advocates for responsible AI/ML have sought ways to draw on the richer causal models of system dynamics to better inform the development of responsible AI/ML. However, a major barrier to advancing this work is the difficulty of bringing together methods rooted in different underlying assumptions (i.e., Dana Meadow's `the unavoidable a priori'). This paper brings system dynamics and structural equation modeling together into a common mathematical framework that can be used to generate systems from distributions, develop methods, and compare results to inform the underlying epistemology of system dynamics for data science and AI/ML applications.
\end{abstract}

\textsuperscript{1} Center for Community Health Integration, Case Western Reserve University\\
\textsuperscript{2} Department of Biomedical Engineering, Case Western Reserve University\\
\textsuperscript{3} Population Health and Equity Research Institute, MetroHealth\\
\textsuperscript{4} Department of Population and Quantitative Health Sciences, Case Western Reserve University

\begin{quote}
\pagebreak

\emph{The direct representation of physical (or social or economics, etc.)
processes leads to the mathematical models whose ``natural'' form can
assume many different shapes.}

\begin{flushright}
(Fred C. Schweppe 1973 p. 21)
\end{flushright}

\emph{In addition to the shared concepts general to all mathematical
modeling, each methodological school also employs its own set of
theories, mathematical techniques, languages, and accepted procedures
for constructing and testing models. Each modeling discipline depends
on unique underlying and often unstated assumptions; that Is, each
modeling method is itself based on a model of how modeling should be
done.}
\end{quote}

\begin{quote}
\emph{These deep, implicit, operating assumptions at the foundation of each
modeling method are sufficently important that they should be
re-examined more often than they actually are.}

\begin{flushright}
(Donella Meadows, 1976, p. 163)
\end{flushright}
\end{quote}

\begin{quote}
\emph{To support a causal inference from observational data, however,
substantial prior knowledge about the mechanisms that generated the
data must be available to justify the necessary assumptions.}

\begin{flushright}
(National Research Council, 2004, p. 8)
\end{flushright}
\end{quote}

\section{Introduction}\label{introduction}

The focus in system dynamics (SD) is on understanding the dynamic behavior of systems from an endogenous or feedback perspective through the use of computer simulation of systems of nonlinear ordinary differential equations (Richardson, 2011). Understanding the dynamics of systems in SD emphasizes developing an explicit explanation of the \emph{structure-behavior relationship} in a system, that is, understanding how the dynamics of a system are generated by a set of interacting nonlinear balancing and reinforcing feedback mechanisms. With advances in implementing formal methods of loop dominance analysis in recent years, this structure-behavior relationship can now be more systematically and rigorously studied by determining the patterns of dominance of feedback mechanisms, which opens up new areas for both theoretical and applied research in understanding the dynamic behavior of complex systems.

System dynamics has also gained interest among data scientists and engineers seeking to advance more responsible AI/ML applications. For example, Ruha Benjamin (2019) sees system dynamics as a potential way to mitigate the biases that appear in AI/ML applications due to incomplete causal knowledge of the societal context by engaging and involving communities in conceptualizing systems using participatory methods. Researchers at Google Research have also been exploring and motivating the application of system dynamics for better understanding societal context of AI/ML (Kuhlberg et al 2023; Martin \& Moore, 2020). Most recently, for example, Martin and Kinney (2024) have focused on demonstrating how the inclusion of feedback loops can help reduce the epistemic uncertainty of AI/ML models. However, these efforts have largely focused on comparing the predictive performance of AI/ML models that include feedback loops as opposed to leveraging the deeper structure-behavior relationship of feedback mechanisms. From a system dynamics perspective, a particular concern has been the difficulty of effectively communicating nonlinear feedback effects involving accumulations to data scientists and statistically oriented fields. Some of the confusion stems from differences in the diagramming conventions, but also more fundamentally in not having a common mathematical framework for comparing methods. While this could be attributed to differences in the underlying assumptions, training, and practices as Donella Meadows described as the ``unavoidable a priori'' in her 1976 conference paper, the various methods \emph{are grounded in mathematics} and hence can be resolved to create a bridge for better multi- and trans-disciplinary collaborations. To address this gap, Meadows called for more frequently revisiting our underlying assumptions and comparing methods.

Such situations are not unique nor new, and efforts to formalize the relationship between different approaches can yield interesting results and applications. For example, there was a long-standing dispute in statistics about the comparability of results between analysis of variance and multiple regression with the resolution eventually leading to the general linear models we rely on today. In another example, path analysis and partial regression analysis ultimately developed into the LISREL model that integrates latent causal modeling and measurement models to form the basis of modern structural equation modeling (Bollen, 1989).

System dynamicists often shy away from learning and using advanced statistical methods such as structural equation modeling. This is unfortunate because methods such as SEM can help identify the longitudinal patterns that form the basis of reference modes, establish measures for intangible or ``soft'' variables, empirically inform some of the causal relationships in our models, and provide additional tools for empirically testing and building confidence in our models and simulation results. This has led to exclusion of intangible variables from models even when they are known to play a critical role in judgement and decision making, reference modes not being appropriately grounded in data, the use of partial regression estimates to inform causal structure, and a tendency to rely on statistical associations to inform model formulation instead of operational thinking skills.

However, acquiring the statistical training to incorporate methods such as structural equation modeling is often outside the scope of most programs, and statistical courses usually begin with a fundamentally different set of assumption leaving it up to the novice student to figure out the underlying relationships. This is usually fraught with conceptual challenges where it may be difficult to distinguish what a novice does not yet know about a method from true conceptual differences. But, the barriers are not limited to the novice learner or student. Even among established experts, the very experience of looking at a system through dynamics and feedback loops or from the perspective of answering a research question given some data leads to deeply ingrained patterns of thought.

While the mental habits may keep us locked in the ``unavoidable a priori'', these methods are mathematical in their description and use. Mathematics is a powerful tool for exploring and translating ideas in ways that break us out of our patterns of thought. The similarity of these methods suggests that bridging these different views is more of a problem with mathematical description than some grand unifying theory or discovery. Hence, this paper is about developing and proposing a mathematical framework that covers both system dynamics and structural equation modeling.

For the field of system dynamics, a mathematical framework in common with related methods would help system dynamicists better integrate advanced statistical tools into system dynamics modeling. For data scientists and engineers, particularly those working to incorporate system dynamics into AI/ML applications, a common mathematical framework would help data scientists more appropriately and effecttively use results from system dynamics studies to advance statistical methods and applications. In this paper, we focus specifically on developing a mathematical framework bridging system dynamics and structural equation modeling (e.g., Bollen, 1989; Gunzler, Perzynksi, and Carle, 2021; Pearl, 2009; Pearl and Mackenzie, 2018).

More generally, causal modeling plays a central role in data science and AI/ML applications that inform the evidence-based interventions, practices, and policies that impact our world from medicine and education to economics and the environment. When the research that informs the development of new products and clinical interventions, for example, excludes theories of systemic causation (Bunge, 1997) as explanations, not only are we likely to misinterpret the effectiveness of interventions and attribute systemic causation (powell,2008) to statistical ``nuisance'' terms, but find ourselves caught in hopeless capability trap of not being able to understand our dynamically complex and changing world at a time when we have the greatest need to understand and apply systems science.

The paper is organized as follows. First, the background provides a brief history of structural equation modeling as it relates to this paper, prior efforts to link SD and SEM, and efforts to develop a unified framework for simulation models. Next, we provide an overview of our approach to developing a general model, which begins by addressing some preliminary differences between SD and SEM. We then develop set of conventions and general mathematical framework to include SEM models. We then provides several examples of SD and SEM models using this general framework, and close with a discussion of some of the limitations, next steps, and implications.

\section{Background}\label{background}

In this section, we provide a brief history of SEM and relevant threads of prior work to develop a general framework for simulation modeling. We then discuss several conceptual barriers that have arguably contributed to the difficulties of bridging SD and SEM research and the unavoidable a prior.

\subsection{Brief history of structural equation modeling}\label{brief-history-of-structural-equation-modeling}

Structural equation modeling (SEM) brings together two main components, a structural latent causal model and a measurement model, from factor analysis and path analysis in statistics. Charles Spearman (1904) is credited with constructing the first factor model, which laid the foundations for measuring latent variables (Spearman, 1904; Tarka, 2018). In his work, he proposed a general ability factor, which he called g, that influences performance on disparate cognitive ability measures (Tarka, 2018). Sewell Wright (1918), a geneticist, was the originator of path analysis (Wright, 1921). In his work, he laid out an analytical strategy for imposing a causal structure to describing causal relationships between measures, and developed path diagrams to graphically depict the relationships between measures in these analyses (Tarka, 2018). Following these origins, SEM has grown due to social scientists and other academics looking to understand the structure of latent phenomena and relationships amongst latent and observed phenomena (Tarka, 2018). As latent variable models have become more complex, the advancement of latent variable software such as MPlus (Muthén \& Muthén, 2012) has played a prominent role in the accessibility of SEM.

\subsection{Efforts to link SD and SEM}\label{efforts-to-link-sd-and-sem}

Prior efforts to link system dynamics modeling involving systems of nonlinear ordinary differential equations with structural equation modeling work by Hovmand (2003) and Hovmand and Chalise (2015), however, these efforts were targeted to address more specific questions and limited to illustrative examples as opposed to a general framework. Related work outside system dynamics has mostly focused on trying to estimate parameters in ordinary differential equation models as dynamical systems using nonlinear regression (e.g., Archontoulis and Miguez, 2015) and structural equation modeling (e.g., Boker and Wenger, 2007). However, these methods largly focus on estimating parameters using the solution to a system of differential equations as opposed to the underlying causal structure. Levine et al's (1992) effort to trace the effects of perturbation through path analysis with feedback loops is an approach similar to the popular MITRE Corporation's Loopy (2021) online tool, however, both do this without accumulations which can lead to misleading inferences on the dynamic behavior of conserved systems.

Schweppe's (1973) work on uncertain dynamic systems provides a general approach uncertain dynamic systems including both frequentist and Bayesian perspectives, static and dynamic systems, discrete and continuous time systems, and linear and nonlinear systems. Schweppe's mathematical framework considers these as a Bayesian linear model with 16 extensions and special cases. However, Schweppe's focus is more on the general estimation problem in uncertain dynamic systems as opposed to seeking to build a bridge between system dynamics and structural equation modeling as methods for representing causal systems.

\subsection{General frameworks for computer simulation}\label{general-frameworks-for-computer-simulation}

There have also been efforts to unify approaches to computer simulation modeling. For example, Zeigler (1976) presented a general mathematical framework for different types of computer simulation modeling, e.g., discrete event simulation, continuous time simulation, etc. And, several software packages integrate different modeling paradigms, the most notable effort in system dynamics being AnyLogic which support system dynamics simulation modeling, discrete event simulation, and agent based modeling within the same integrated platform. These efforts focus mostly on the implementation of computer simulation as an applied mathematics or computer science problem and make contributions to the development of simulation software, but do not provide the general mathematical framework for understanding systems using system dynamics or structural equation modeling. That is, these efforts help address some of the divisions between approaches to computational modeling, but not the larger divide between computational modeling and statistical modeling.

\subsection{Differences between SD and SEM}\label{differences-between-sd-and-sem}

Our approach is to develop a common mathematical framework for bridging system dynamics (SD) modeling as a method with structural equation modeling (SEM). To do this, we first deal with conceptual barriers in interdisciplinary collaborations between SD and SEM. Table 1 provides a summary of conceptual differences between SD and SEM features, which will elaborated on in the next section as preliminary concepts before presenting our general framework. These differences are issues we will need to address in developing our general framework.

\textbf{Table 1.} Summary of conceptual differences between SD and SEM

\begin{longtable}[]{@{}
  >{\raggedright\arraybackslash}p{(\columnwidth - 4\tabcolsep) * \real{0.3056}}
  >{\raggedright\arraybackslash}p{(\columnwidth - 4\tabcolsep) * \real{0.3472}}
  >{\raggedright\arraybackslash}p{(\columnwidth - 4\tabcolsep) * \real{0.3472}}@{}}
\toprule\noalign{}
\begin{minipage}[b]{\linewidth}\raggedright
Feature
\end{minipage} & \begin{minipage}[b]{\linewidth}\raggedright
System Dynamics
\end{minipage} & \begin{minipage}[b]{\linewidth}\raggedright
Structural Equation Modeling
\end{minipage} \\
\midrule\noalign{}
\endhead
\bottomrule\noalign{}
\endlastfoot
Goals & Focus is on explaining the observed behavior of a system & Focus is on explaining the observed covariance matrix \\
Problem type & Inverse problem of finding a nonlinear feedback system and parameters & Forward problem of estimating effects and prediction \\
Diagramming conventions & Causal loop diagram and stock and flow diagram & Path diagram with indicators \\
Equations & (1) system of ordinary differential equations and (2) their solution, usually simulated & (1) system of linear equations and (2) their implied covariance matrix \\
Dynamics & Dynamic and static subsystems & Static system \\
Continuity & Continuous time and value & Discrete time and continuous values \\
Fit with observations & Standard measures of fit (e.g., \(MSA\), \(RMSEA\), \(R^2\)) between the global fit of simulated behavior and observed behavior over time, decomposition of error using Theil inequality statistics (e.g., \(U^M\), \(U^S\), and \(U^C\)) & Both measures of global fit (e.g., \(\chi^2\), \(RMSEA\), \(GFI\), \(AGFI\), \(R^2\)) and local fit (e.g., modification indices, \(d\)-separation, \(trek\)-separation) between the implied covariance matrix and observed covariance matrix and measures \\
Specification & Inherently under-determined & Identified or over-determined \\
\end{longtable}

It is important to note, first, that the way we resolve these is to develop a common mathematical framework as opposed to the most parsimonious way of describing either method alone. Second, we focus on a common framework that generally covers the usual models from SD and SEM as opposed to a creative way that a method can be embedded within another. For example, system dynamics simulation software can be used to simulate a discrete set of observations and generate results that appear similar to the observed covariance matrices that form the basis of structural equation modeling. While such tricks may satisfy a specific application, they do not provide a general mathematical framework nor lead to a better understanding of how the two approaches are related.

\subsubsection{Goals}\label{goals}

In SD, the primary goal is to develop an explanation for a dynamic behavior pattern of interest that is usually represented as a behavior over time graph and often referred to as the reference mode. In contrast, the primary goal of SEM is to estimate the effects and predict an observed covariance matrix from the implied covariance matrix of a causal model.

Differences between the goals of the two methods arguably sit within a broader context of concerns about scientific methods, especially as this applies to social systems where the ontology of socially constructed systems (Berger and Luckmann, 1966) requires a scientific strategy that places a greater emphasis on theory appraisal before committing to statistical tests of propositions (Lakatos, 1970; Meehl, 1990).

\subsubsection{Problem type}\label{problem-type}

SD is essentially about solving the \emph{inverse problem} of finding a plausible system of feedback mechanisms that can generate the dynamic behavior pattern being studied and the associated parameters of a model whereas SEM is primarily about solving the \emph{forward problem} of predicting the effects from a given model.

The distinction between solving the forward problem versus the inverse problem is nontrivial. In the forward problem, the goal is to predict a response variable based on a model of \emph{predictor} variables. For example, in \(y=a + b \cdot x\), \(y\) is the response variable, and \(x\) is the predictor variable with \(a\) and \(b\) constants. This can be simplified into a functional description as \(y=f(x,a,b)\). When focused on inverse problems, the goal is fundamentally different where we want to find the function \(f\) where \(y=f(x,a,b)\). That is, we are less concerned with how well our identified function \(f\) that maps \((x,a,b)\) onto \(y\) fits, and more interested in \emph{which} functions, if any, could do this.

Confusion often arises in interdisciplinary conversations trying to understand SD results when people ask how the parameters or strength of the effects in causal relationships are determined. From the perspective of SD, there are lots of ways to assign a value to a parameter or causal relationship in a system dynamics model including grounding parameters empirically using a range of mixed methods. (i.e., from Delphi methods with experts to numerical estimates derived from meta-analysis) and stating Bayesian priors. However, the question is often misplaced because SD is focused on solving the inverse problem of finding plausible systems of differential equations and parameters that could explain the dynamics of the reference mode, not the forward problem of estimating the parameters and the effects of a model, which is often the primary concern in areas such epidemiology and biostatistics.

What we ultimately want to do is show these two methods of solving the inverse problem and forward problem are complementary, not an either/or, and understanding this is essential for gaining a better grasp of the complexity of social systems for developing and sustaining a strong progressive program of research (Lakatos, 1970).

\subsubsection{Diagramming conventions}\label{diagramming-conventions}

System dynamics is essentially about finding a set of causal feedback relationships that can generate a given behavior pattern of interest, that is, solving the inverse problem as opposed to the forward problem of predicting what will happen given a set of causal relationships. System dynamics models are typically represented as either a causal loop diagram or a stock and flow diagram (Figures 1a and 1b) whereas structural equation models use path diagrams (Figure 1c). Although appearing similar, the conventions differ. For example, system dynamics models focus on identifying the set of balancing and reinforcing loops with loop labels using a B prefix for balancing loops and R prefix for reinforcing loops, whereas structural equation models generally do not even include explicit labels for feedback when they appear as cycles or non-recursive relationships as shown in Figure 1c.

More significant is the fact that structural equation models make an explicit distinction between (1) latent or unobserved variables as circles or ovals and (2) indicators or observed variables as rectangles (Figure 1c). In system dynamics, the main distinction is between stocks and flows, as shown in a stock and flow diagram (Figure 1b) where rectangles represent accumulations or stocks, and circles represent converters that include rates of change or flows (shown in Figure 1b with a value), endogenous variables, and exogenous variables and constant parameters. Causal loop diagrams generally do not make a visual distinction between stocks and flows and therefore drop the use of circles and rectangles altogether.

System dynamics explicitly call out exogenous parameters in a model, e.g., \(\beta1\) and \(\beta2\) in Figures 1a and 1b, however, structural equation models generally do not because these are to be estimated by fitting the specified model to data. Meanwhile, standard structural equation models will generally include not only the indicators (\(X1\), \(X2\), \(X3\), and \(X4\)) and their corresponding error terms in the measurement model (\(\varepsilon1\),\(\varepsilon2\), \(\varepsilon3\), and \(\varepsilon4\)) (Figure 1c).\footnote{Note that extensions of the standard SEM model allow for error terms to be correlated.}

\begin{figure}
\centering
\includegraphics{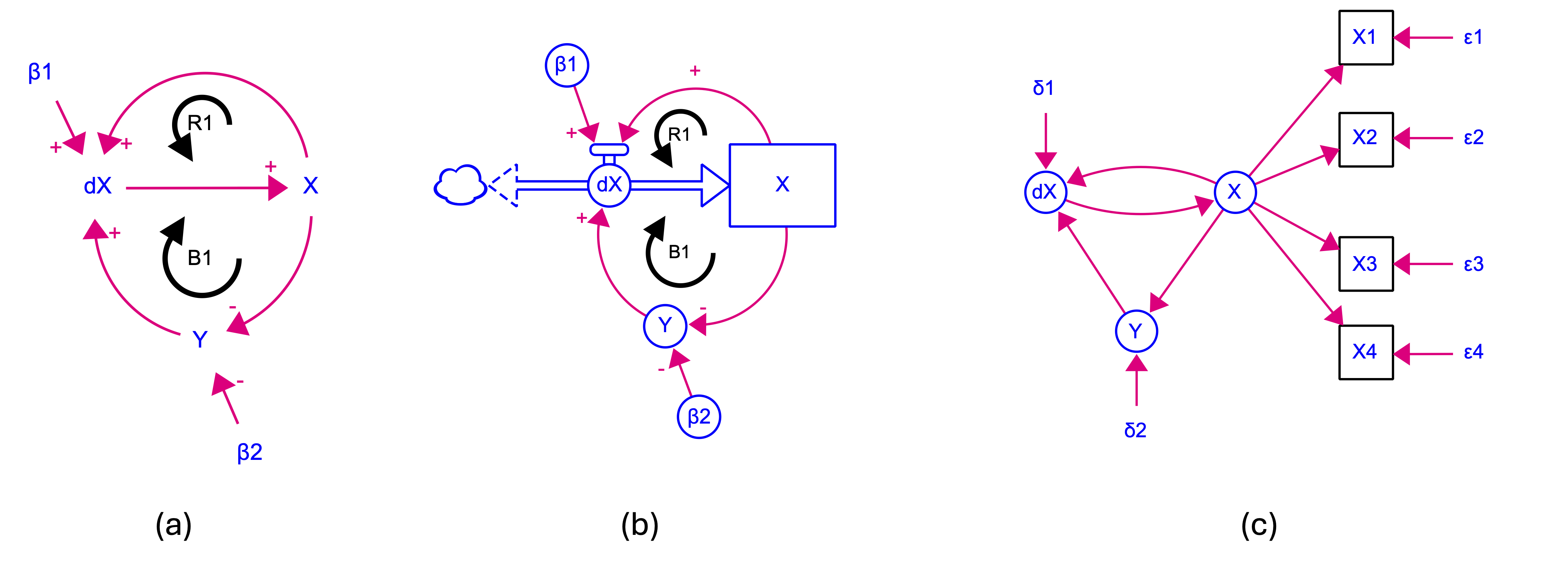}
\caption{System dynamics diagraming conventions of (a) causal loop diagram, (b) stock and flow diagram, and (c) a path diagram from structural equation modeling}
\end{figure}

Adding to the confusion is that the relationship between causal loop diagrams (CLDs) and stock and flow diagrams (SFDs) in system dynamics is often ambiguous when going from a CLD to an SFD, but exact when going from an SFD to a CLD because CLDs are often conceptualized without explicitly identifying the accumulations and their corresponding flows. Figure 1a comes from drawing the CLD \emph{based on} the SFD shown in Figure 1b. However, there are often numerous ways to translate a CLD into a SFD (Figure 1a being an exception by explicitly identifying the rate of change as \(dX\) for the variable \(X\) which would imply that \(X\) is an accumulation or stock variable).

The ambiguity of CLDs concerning their translation into SFDs leads some to caution against their use in system dynamics because they contribute to errors in conceptualizing systems and interpreting the behavior of systems (Richardson, 1997). Nonetheless, CLDs remain a popular choice in systems thinking because the diagramming conventions are often seen as easier to learn and use, especially when the goal is not to develop a formal system dynamics computer simulation model. Moreover, their similarity to the causal structure of directed acyclic graphs (DAGs) and latent causal structures in path diagrams have encouraged statisticians and more recently data scientists to view CLDs as an extension of DAGs, usually without recognizing some of their limitations and appreciating the significance of accumulations in feedback systems.

We will need to address these differences in formulating a general SD-SEM framework by defining a set of diagramming conventions that highlight the homomorphism of symbols between the two approaches. To do this, we will ultimately show an SD and SEM version of each system for each of the subsequent examples in hopes that this will enable readers trained in one approach to follow the argument and see the connection to the other approach.

\subsubsection{Equations}\label{equations}

System dynamics and structural equation modeling deal with two entirely different sets of equations. System dynamics builds on a framework of nonlinear differential equations, which have an causal system represented by a system of differential equations, and an analytic or numerically approximated solution using computer simulation. Structural equation modeling uses a system of linear equations to represent the causal system and measurement model, which can be generalized to include nonlinear interaction terms, and an implied covariance matrix. Confusions arise between structural equation modeling and system dynamics modeling because people tend to either (1) associate the causal system of differential equations in SD with the linear equations representing the latent causal structure in SEM, or (2) associate the solution to the system of differential equations in SD with implied covariance matrix in SEM. The next two subsections aim to make this distinction more explicit.

\paragraph{SD - systems of nonlinear ordinary differential equations}\label{sd---systems-of-nonlinear-ordinary-differential-equations}

SD models consist of a system of nonlinear ordinary differential equations (ODEs), which are specified as a set of integral equations representing causal relationships in the system:\footnote{The symbol \(\int_{t_0}^{t}\) is the integral from \(t_0\) to time \(t\). To avoid confusion between the time \(t\) that we are integrating to and the time we are integrating over between \(t_0\) to time \(t\), we follow the convention of \(u\) and \(du\) as dummy variables for \(t\).}

\[
\mathbf{x(t)} = \int_{t_0}^{t}f(\mathbf{x(u)},\mathbf{c})du + \mathbf{x_{t_0}}
\]

where \(\boldsymbol{x(t)}\) is a vector of state variables or stocks, \(\boldsymbol{x(t)} = (x_1(t),...,x_k(t))\) at time \(t\), where we use lowercase Roman bold font for vectors. The function \(f\) is a vector of rates of change or flows in the system, which is a function of the vector of stock variables, \(\boldsymbol{x(t)}\), and a vector of constants \(\boldsymbol{c} = (c_1, ..., c_j)\). The initial values of the stocks at \(t_0\) are defined by the vector \(\boldsymbol{x_{t_0}} = (x_1(t_0),...,x_k(t_0))\).

When a system of integral equations is used to describe a causal system and its behavior, there are essentially two different sets of equations involved: (a) the system of differential equations representing the causal system as rate equations, and (b) the solution to the system of integral equations as shown below.

\[
\boldsymbol{x(t)} = \underbrace{\int_{t_0}^{t}\overbrace{f(\boldsymbol{x(u)}, \boldsymbol{c})}^{(a) \,causal \, system}du + \boldsymbol{x(t_0)}}_{(b) \, solution}
\]

For linear systems of ordinary differential equations, we can generally find a solution to a given system. For example, in a population model where births, \(births(t)\) are proportional to the size of the population,\(pop(t)\), we have, \(births(t) = c \cdot pop(t)\) where \(c\) represents a constant, i.e., fractional birth rate. This is the \emph{causal system} for this simple linear model, which has the following solution:

\[
pop(t) = \int_{t_0}^{t}births(u)du + pop(t_0) = e^{c \cdot t} + pop(t_0).
\]

Hence,

\[
pop(t)/dt = births(t)=  c \cdot pop(t), \\ \tag{1}
\] \[
pop(t) = e^{c \cdot t} + pop(t_0). \\ \tag{2}
\]

Equation (1) is the causal system relating population size to birth rate, and equation (2) is the solution to (1) where population, \(pop(t)\), is the behavior of the system over time \(t\) and a function of the constant \(c\) and initial population \(pop(t_0)\) at \(t_0\). If we can analytically solve (1) to find (2), then we can use data to estimate the parameter \(c\) with a nonlinear regression and draw inferences about how a change in \(c\) or the initial value \(pop(t_0)\) affects the behavior over time of \(pop(t)\).

However, methods for finding analytic solutions of the form (2) are generally only available for systems of \emph{linear} ordinary differential equations and special cases of nonlinear systems of ordinary differential equations. System dynamics is primarily concerned with systems of \emph{nonlinear} ordinary differential equations where a common approach is to solve the system of equations through numerical. integration methods. That is, in system dynamics, we find a numerical approximation of the solution (2) by \emph{simulating} equation (1) over time. On the occassions when an analytic solution is available, we can compare the analytic solution (2), which is exact, against our numerically approximated solution from simulation to evaluate the performance of the numerical approximation.

This distinction becomes important when considering the differences because there are approaches in SEM to modeling dynamical systems that rely on estimating coefficients of an underlying causal system from data for where the analytic solutions are already known. This usually entails assuming a simple mathematical causal description of a system based on known mathematical models from biology or physics, e.g., population growth, oscillating pendulum, and seeking find mechanisms in the real system of interest that correspond to simple biological or physical mechanisms. There can be many situations where this is convenient and makes sense scientifically, but it is fundamentally different from the exercise in system dynamics of building up a causal system based on operational knowledge of the system and other known or hypothesized mechanisms (e.g., from scientific literature, interviews, group model building).

\paragraph{SEM - systems of linear equations}\label{sem---systems-of-linear-equations}

Stuctural equation modeling builds on the idea of representing a system as a set of relationships between a vector endogenous latent variables \(\boldsymbol{\eta} = (\eta_1,...,\eta_m)\) measured by a vector observed variables \(\boldsymbol{y}=(y_1,...,y_p)\),, and a vector of exogenous latent variables \(\boldsymbol{\xi}=(\xi_1,...,\xi_n)\) measured by a vector of observed variables \(\boldsymbol{x}=(x_1,...,x_q)\) (Gunzler, Perzynksi, Carle, 2021). The relationships are described in the LISREL matrix form as \emph{structural model} and \emph{measurement model}.

The latent causal model is defined as,

\[
\boldsymbol{\eta}= \boldsymbol{\alpha}_{\eta} + \boldsymbol{\mathrm{B}\eta} + \boldsymbol{\Gamma\mathbf{\xi}} + \boldsymbol{\zeta} \tag{3}
\]

where \(\mathbf{\alpha}_{\eta}=(\alpha_1,...,\alpha_m)\), \(\mathrm{B}\) is a \(m \times m\) matrix of slopes relating the endogenous latent variables, and \(\Gamma\) is a \(m \times n\) matrix of slopes related endogenous to exogenous latent variables, and \(\mathbf{\zeta}\) is a vector of random error terms, \(\mathbf{\zeta}=(\zeta_1, ..., \zeta_m)\).

The measurement model is then defined as,

\[
\begin{split}
\boldsymbol{y} & = \boldsymbol{\nu}_y + \boldsymbol{\Lambda}_y\boldsymbol{\eta} + \boldsymbol{\epsilon} \\
\boldsymbol{x} & = \boldsymbol{\nu}_x + \boldsymbol{\Lambda}_x\boldsymbol{\xi} +  \boldsymbol{\delta}
\end{split} \tag{4}
\]

where \(\boldsymbol{\nu}_y\) and \(\boldsymbol{\nu}_x\) are vectors representing intercepts for \(\boldsymbol{y}\) and \(\boldsymbol{x}\) respectively, \(\boldsymbol{\Lambda}_y\) and \(\boldsymbol{\Lambda}_x\) represent matrices of slopes or factor loadings for \(\boldsymbol{y}\) and \(\boldsymbol{x}\) respectively, and \(\boldsymbol{\epsilon}\) and \(\boldsymbol{\delta}\) are vectors representing measurement errors for \(\boldsymbol{y}\) and \(\boldsymbol{x}\) respectively.

\subparagraph{Implied covariance matrix}\label{implied-covariance-matrix}

Matrix equations (3) and (4) form the basis for calculating an implied covariance matrix \(\boldsymbol{\sum(S)}\) from parameters \(\boldsymbol{S}\) that can be compared against the observed covariance matrix \(\boldsymbol{\sum}\). Restrictions arise in SEM from the assumptions required to calculate the implied covariance matrix from (3) and (4). Extensions to (3) and (4) allow for the inclusion of nonlinear terms such as a polynomials and interactions between two latent variables.

\subsubsection{Dynamics}\label{dynamics}

The term `dynamic' takes on different meanings in system dynamics and structural equation modeling. In structural equation modeling, particularly \emph{dynamic} structural equation modeling, the term `dynamic' means change over time, e.g., in longitudinal and time series patterns. However, in system dynamics, the term `dynamic' refers not specifically to the behavior of a system changing over time, but the formal mathematical distinction between a static versus a dynamic system (see, for example, Palm 1983). In a \emph{static system}, the output of a system \(y(t)\) is strictly a function of the input \(x(t)\), i.e., \(y(t)=f(x(t))\). In a \emph{dynamic system}, the output is a function of both the input \emph{and} the state of the system, i.e., \(y(t)=f(x(t),y(t))\). That is, what differentiates a static and dynamic system is not whether the output is changing over time, but whether how the output changes with the input. Consider an example of turning a light on and off, one and off. As soon as the input stabilizes, the output is stable (the light is either on or off). The light switch and light in this example represent a static system.

That is not true in a dynamic system. Consider an example of central heating and cooling in a building where one increases the desired temperature relative to some outside temperature with the expectation that the indoor temperature will rise to match the desired temperature. Once the desired temperature is set and assuming the outdoor temperature also remains constant, the indoor temperature will continue to change toward the desired temperature even though the inputs are stable.

The distinction between static and dynamic systems is something we will leverage in formulating our general SD-SEM framework and use to show how the general SEM framework maps into the static portion of a system dynamics model.

\subsubsection{Continuity}\label{continuity}

Both system dynamics and structural equation modeling generally assume that the underlying variables of the causal system are continuous, although there are ways to extend both to allow for discrete values. However, system dynamics and structural equation modeling differ in how variables are viewed over time.

Structural equation modeling builds on a set of finite observations, which are discrete. System dynamics models simulate a continuous system over time, which is in contrast to discrete event simulation and agent based modeling, which simulate a system in discrete events. The continuous time perspective often gets confused with the discrete time approximation of a continuous system, so that people often make the mistake of assuming that the discrete time-step (\(DT\) or \(\Delta T\)) of numerical integration used to calculate an approximate solution to the continuous time system should be the same as the observation times. That is, if there 10 observations of the real system, then the system dynamics model should be simulated in 10 time steps.

This confusion gets compounded when longitudinal observations are recorded and analyzed by their position in a longitudinal sequence (i.e., \(y_1, y_2, ..., y_k\)) as opposed to their actual time of data collection (i.e., \(y(t_1), y(t_2), ..., y(t_k)\)). Varying differences between the actual time of observation, i.e., \(t_2 - t_1, t_3-t_2, ..., t_k-t_{k-1}\), can lead to distortions in the shape or behavior modes of a curve, especially when the sampling period is large relative to the time constants in a system dynamics model.

This has implications for how we handle measurement error in developing a general framework because we need to consider both the measurement error associated with the value and time when bringing discrete observations into a measurement model for system dynamics.

\subsubsection{Measures of fit}\label{measures-of-fit}

Approaches to measures of fit between models and observations differ considerably between system dynamics and structural equation modeling both in their respective practices and development. Whereas structural equation modeling has placed a strong emphasis on developing and critically using both global and local measures of fit along with standard reporting guidelines that emphasize multiple measures versus relying on a single measure, system dynamics has not. For example, despite an insistence by Sterman (2000) to provide an assessment of fit and decompose the error using Theil inequality statistics, most published system dynamics models still do not report as a matter of standard practice their fit with observed reference modes.

While this might be attributed to differences between the underlying goal of modeling in system dynamics and structural equation modeling and greater reliance on measures of fit in structural equation modeling for selecting and refining models, there is essentially no excuse in system dynamics for not providing model assessments of fit even if these are known to be weak confidence building tests.

We will return to measures in fit as the general model we propose allows for drawing on work in structural equation modeling and specifically, ways to leverage the newer methods in system dynamics for predicting link and loop scores as a basis for formulating structural equation models that solve the forward problem of prediction.

\subsubsection{Specification}\label{specification}

SD models are inherently under-determined, meaning that no set of observations will uniquely identify the system dynamics model and parameters. There is, in fact, an infinite set of models and parameters to consider. However, SD does not rely on numeric data alone to identify a model and recognizes that in social systems, people are also actors and observers of systems and hence have subjective qualitative knowledge about the structure of the system and decision rules within the bounded rationality of experience (e.g., Forrester, 1980; Sterman, 2000). SD models in this sense are not uniquely identified in the numeric sense, but given that the goal is to solve the inverse problem, that is also not critical.

The focus on solving the forward problem in SEM means the model needs to be identifiable, and hence models in SEM need to be identifiable in their specification, which emphasizes developing and using heuristics such as the minimum number of indicators per latent variables and concerns about the number of observations where models in SEM may be uniquely determined or overdetermined.

\section{General SD-SEM framework}\label{general-sd-sem-framework}

This section presents the overall framework from the perspective of the system dynamics modeling process where we start with a reference mode, followed by the diagramming conventions we will use for our framework, and then the general mathematical form of the proposed SD-SEM framework with some examples.

\subsection{Reference mode}\label{reference-mode}

The concept of a reference mode is unique to system dynamics and often misunderstood outside the field along with the significance of defining a reference mode for understanding systems. A reference mode is a formal statement of the focal dynamic problem or issue of interest which can be stated in words, but more typically as a graph of system behavior over time (e.g., Richardson and Pugh, 1986).

Reference modes can consist of one or more variables over time, but typically only a few variables at most as the goal is to focus the problem or issue down to a few variables. This effort to focus the problem on a few variables over a specific time horizon is a form of problem structuring, and something that distinguishes system dynamics from more general systems thinking and causal mapping that often involve many variables and feedback loops without consideration of a reference mode.

Hence, central to modeling a feedback system in defining a reference mode is the selection of the time horizon of the reference mode because this essentially limits which feedback loops and accumulations are going to be relevant for the investigation. In the real system, dynamic phenomena are occurring simultaneously at multiple scales from the smallest interval of time of \(10^{-43}\) seconds (i.e., Planck time) to billions of years (i.e., Eons). However, defining the time horizon in a reference mode gives us a temporal lens to view view the dynamics and relevant feedback loops and accumulations.

It is important to note that reference modes as problem structuring are \emph{idealized abstractions} of the system behavior that is of interest and hence not identical with the longitudinal or time series data, nor restricted to what data are available. That said, reference modes \emph{should} be empirically grounded. Building a model to solve a dynamic problem that is not supported by data risks solving fictional problems. There are too many real problems in the world to work on that can benefit from system dynamics insights to be chasing problems that are not grounded in reality.

The type of data that might be used to support a reference mode can range from qualitative graphs over time elicited through group model building workshops to longitudinal data and intensive time series numeric data. A key methodological decision in system dynamics is selecting the appropriate time horizon and variables for defining the reference mode, and hence questions arise about the implications of choosing a different time horizon or having more variables as part of the reference mode. Other issues arise as well in how we empirically ground a reference mode, especially when using subjective qualitative recalling trends over time or rely on a limited number observations over time with measurement error.

\begin{figure}

{\centering \includegraphics[width=4in]{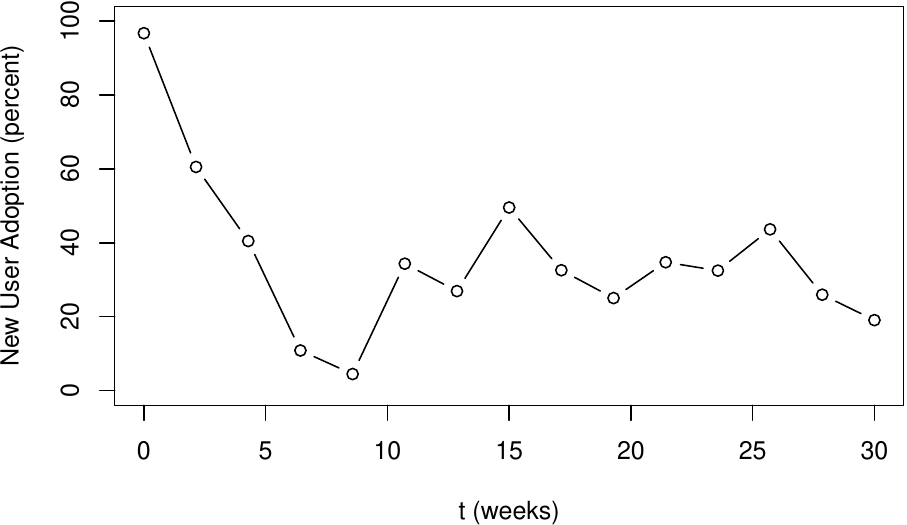} 

}

\caption{Observed behavior pattern over time $b(t)$}\label{fig:data-ref}
\end{figure}

\begin{figure}

{\centering \includegraphics[width=4in]{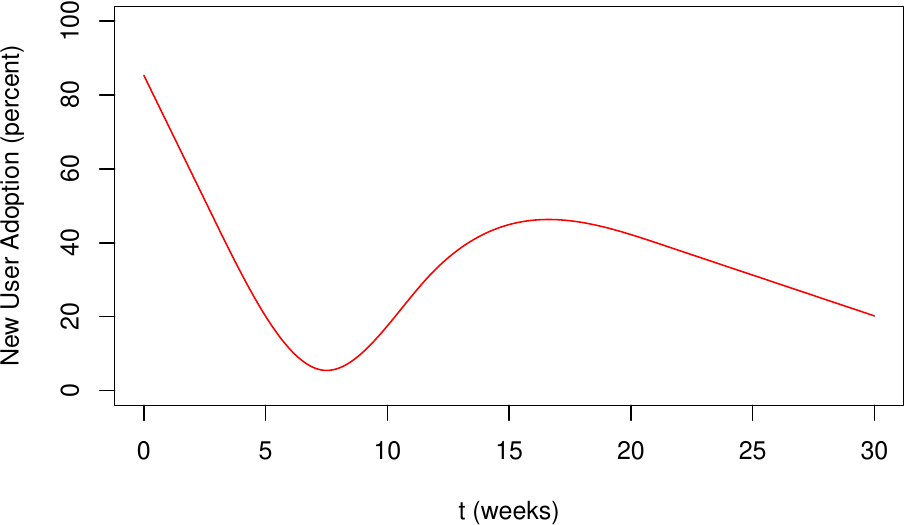} 

}

\caption{Reference mode $r(t)$ for $New User Adoption$}\label{fig:ref-mode}
\end{figure}

In our framework, we distinguish the observed data over time \(\mathbf{b(t)}\) from our reference mode \(\mathbf{r(t)}\) where the claim is that \(\mathbf{r(t)} \approx \mathbf{b(t)}\). If we consider a hypothetical example of a problem focusing on new user adoption over a time horizon of \(h=30\) weeks from our initial time of \(t_I=0\) to our final time of \(t_F\), Figure \ref{fig:data-ref} is observed behavior of the system \(b(t)\) and Figure \ref{fig:ref-mode} reference mode \(r(t)\) based on \(b(t)\). Note that we could have alternative definitions of our dynamic problem even on this time scale, focusing on the overall trend instead of including the damped oscillation or focusing on only the oscillation as opposed to the declining trend. How we choose to define a problem is not visible to others for review, critique, and later communicating the insights to others unless we have made our reference mode explicit and distinguished this from the data we use to define the reference mode.

\subsection{Diagramming conventions}\label{diagramming-conventions-1}

Figure \ref{diag-conventions} shows the two diagramming conventions we will adopt for depicting the system of causal relationships and observed variables as SD and SEM diagrams that are equivalent in the sense that every symbol in one diagram has an exact correspondence in the other and \emph{vice versa}. This should not be interpreted to imply the claim that SD and SEM are mathematically equivalent, i.e., that for every SD model there is a mathematically equivalent SEM model and for every SEM model there is a mathematically equivalent SD model. Instead, this diagrammatic equivalence is a precondition to any claims of mathematical equivalence between SD and SEM. That is, if SD and SEM are mathematically equivalent, \emph{then} there would be at least one isomorphism between SD and SEM diagramming conventions. Our claim here is that this pair of diagramming conventions is sufficient to meet this precondition.

\subsubsection{System dynamics}\label{system-dynamics}

\begin{figure}
\centering
\includegraphics{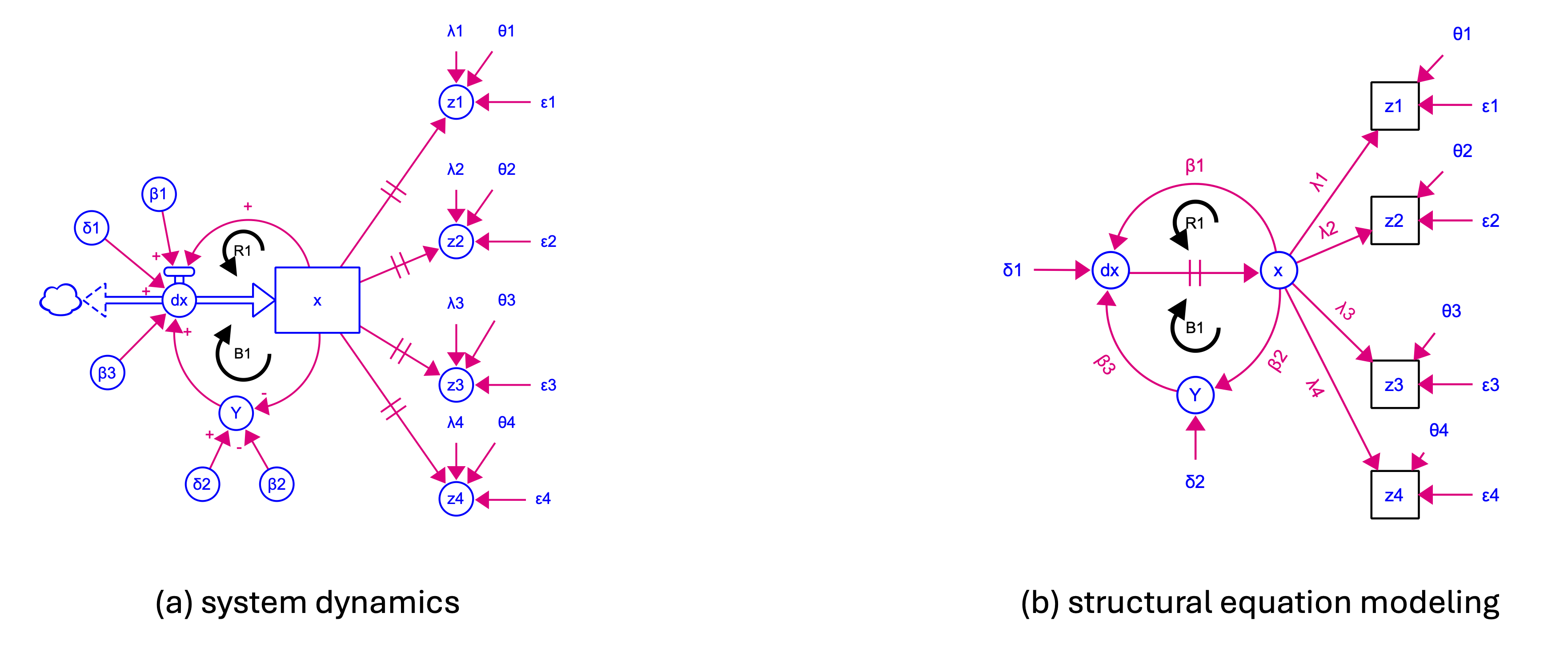}
\caption{Diagramming conventions for SD-SEM framework\label{diag-conventions}}
\end{figure}

Figure \ref{diag-conventions}a shows the SD conventions we will use. Circles will be used to represent converters or auxiliary variables. These variables we simulate over time. Loop labels (e.g., R1 and B1) refer to the strength of the loop, e.g., their normalized loop score. Boxes represent accumulations as stocks. The double lines (e.g., defining \(dx\)) defines \(dx\) as a transition or flow into and out of the stock \(x\). A double line across a causal link indicates a delay. All variables in circles and boxes (converters, flows, and stocks) can in principle be measured or observed from a system dynamics perspective. We will use \(\delta\) prefixes to represent exogenous disturbance terms (e.g., step functions, pulse functions, random noise) and \(\beta\) prefixes to represent parameters (e.g., time constants, fractional rates, proportions).

The symbols \(\lambda\), \(\theta\) and \(\epsilon\) take on special meaning as unknown parameters in our measurement model. They are included in our model and simulated, but are not variables that would be considered exogenous or endogenous in the system dynamics sense. More specifically, drawing on structural equation modeling, the true values of these parameters are always unknown. We may be able to estimate them once we have specified and identified our model, but we do not know their true value. Moreover, we can make assumptions about whether or not we will allow for these terms to be correlated by using the curved double arrow convention from SEM measurement models.

Figure \ref{diag-conventions}a shows a measurement model with \(z1\), \(z2\), \(z3\) and \(z4\) being the observed indicators of \(x\) with \(\lambda1\), \(\lambda2\), \(\lambda3\) and \(\lambda4\) representing the correlation between the indicators and \(x\), \(\theta1\), \(\theta2\), \(\theta3\) and \(\theta4\) the temporal lag or delay between indicators and \(x\), and \(\epsilon1\), \(\epsilon2\), \(\epsilon3\) and \(\epsilon4\) the measurement errors for each indicator where in the standard SEM model, the measurement errors are uncorrelated, i.e., they satisfy the assumption of being independent and identically distributed.

\subsubsection{Structural equation modeling}\label{structural-equation-modeling}

Figure \ref{diag-conventions}b shows the equivalent SEM conventions that we will use. Circles represent latent variables and boxes represent observed variables or indicators. The terms \(\beta1\), \(\beta2\), and \(\beta3\) represent the parameters being estimated the define the strength of the hypothesized causal relationship between latent variables. The terms \(\delta1\) and \(\delta1\) represent exogenous disturbance terms. \(\lambda1\), \(\lambda2\), \(\lambda3\) and \(\lambda4\) are the correlations between the indicators and latent variable. We include \(\theta1\), \(\theta2\), \(\theta3\) and \(\theta4\), which are not part of a standard measurement model, to represent the lags between the latent variable and indicators. \(\epsilon1\), \(\epsilon2\), \(\epsilon3\) and \(\epsilon4\) take their usual meaning in SEM as measurement error terms.

SEM does not have a convention for representing causal relationships involving accumulations of rates of change, which has led to some confusion in trying to estimate causal effects that affect state variables as direct effects other than a rate of change. This stems from a long-standing confusion about the direction of causation in ordinary differential equations and their equivalent integral equations going back to at least to 1894 when Heinrich Hertz (2003) identified the issue in \emph{The Principles of Mechanics Presented in a New Form}, a point that has been reiterated by Forrester in talks and papers including \emph{Principles of System Dynamics} (1990) cautioning against the misinterpretation of direction of causality from taking a derivative.

However, the nature of causation is that the rate of change is \emph{the direct effect} that causes the state variable to change, not the other way around. This is a mathematical relationship, not a statistical relationship. The \emph{direct causes} of the level of water in a bathtub are from the faucet(s) and drain(s). This is what rates of change ``do'' mathematically, they directly change the stocks. This does not mean that the stock and other variables cannot affect the the rates or flows, but it is always \emph{indirectly} through a rate of flow variable.

What is missing from the SEM diagramming conventions is a way to convey a \emph{mathematical} relationship of accumulations from rates of change or flows. In stating this is a mathematical relationship, we mean this is not a statistical relationship in the way that \(\beta1\), \(\beta2\), \(\beta3\), or \(\beta4\) are estimated from data, but a mathematical relationship that follows from the accumulation or integration of a rate of change over time.

To address this, we use the SD convention of delay symbol or double line across a causal link to indicate accumulation. Hence, the causal link from \(dx\) to \(x\) indicates that \(x\) \emph{accumulates} or \emph{integrates} \(dx\) over time. There is no parameter to be estimated from this causal relationship because the causality of integration is mathematical, i.e., it follows from our statement of a derivative, rates of change, or flows with respect to a state variable or stock.

\subsection{System of equations}\label{system-of-equations}

Our approach for the general framework is to decompose the model of a system into three sets of equations or subsystems: the dynamic subsystem, the static subsystem, and the measurement subsystem. The dynamic subsystem describes the rate equations as a matrix of static variables. The static subsystem describes the static variables as a matrix of linear and interaction terms between the stocks or state variables and static variables. The measurement subsystem describes the indicators or observed variables as a matrix of linear combinations of dynamic and state variables of the model.

In our general framework, we consider a system over time \(t\) from the start or initial time \(t_I\) to the end or final time \(t_F\) with a time horizon of \(t_H = t_F - t_I\), and a model with \(m\) latent state variables or stocks represented as a column vector \(\boldsymbol{x}\), \(n\) latent static variables or auxiliaries as a column vector \(\boldsymbol{y}\), and \(p\) indicators or observed variables as a column vector \(\boldsymbol{z}\) with \(q\) observations over \([t_I,t_F]\) in the observed data matrix or frame \(\boldsymbol{O}\),

\[
\mathbf{x} = \begin{pmatrix}x_1 \\ \vdots \\ x_m\end{pmatrix}, \,
\mathbf{y} = \begin{pmatrix}y_1 \\ \vdots \\ y_n\end{pmatrix}, \,
\mathbf{z} = \begin{pmatrix}z_1 \\ \vdots \\ z_p\end{pmatrix}, \,
\mathrm{and}\, \mathbf{O} = \begin{pmatrix}
                        z_{1,1} & \cdots & z_{1,q} \\
                        \vdots  & &  \vdots \\
                        z_{p,1} & \cdots & z_{p,q}
             \end{pmatrix}. 
\]

Table 2 provides an overview of the symbols that will be used in the framework along with their meaning.

\textbf{Table 2.} Symbols and their meaning in the framework

\begin{longtable}[]{@{}
  >{\raggedright\arraybackslash}p{(\columnwidth - 2\tabcolsep) * \real{0.2812}}
  >{\raggedright\arraybackslash}p{(\columnwidth - 2\tabcolsep) * \real{0.7188}}@{}}
\toprule\noalign{}
\begin{minipage}[b]{\linewidth}\raggedright
Symbol
\end{minipage} & \begin{minipage}[b]{\linewidth}\raggedright
Meaning
\end{minipage} \\
\midrule\noalign{}
\endhead
\bottomrule\noalign{}
\endlastfoot
\(t_I\) & Start of time horizon \\
\(t_F\) & End of time horizon \\
\(x(t)\) or \(x\) & Latent dynamic state variable or stock \\
\(y(t)\) or \(y\) & Latent static variable \\
\(z(t)\) or \(z\) & Observed variable or indicator \\
\(m\) & Number of latent dynamic state variables or stocks \\
\(n\) & Number of latent static variables \\
\(p\) & Number of observed variables or indicators \\
\(q\) & Number of observations over \([t_I,t_F]\) \\
\(t_H\) & Time horizon, \(t_H = t_F - t_I\) \\
\(t\) & Specific time in \([t_I,t_F]\) \\
\(\boldsymbol{b(t)}\) & Vector of variables describing behavior of real system at \(t \in [t_I,t_F]\) \\
\(\boldsymbol{o(t)}\) & Vector of variables describing observed behavior of real system at \(t \in [t_I,t_F]\) \\
\(\boldsymbol{r(t)}\) & Vector of variables describing reference mode of interest at \(t \in [t_I,t_F]\) \\
\(\boldsymbol{s(t)}\) & Vector of variables describing simulated values at \(t \in [t_I,t_F]\) \\
\(\beta\) & Model parameter \\
\(\gamma\) & Exponent \\
\(\lambda\) & Association or factor loading between latent variable and observed or indicator variable \\
\(\theta\) & Mean time delay or lag of observed or indicator variable \\
\(\delta\) & Disturbance term or exogenous perturbation \\
\(\epsilon\) & Measurement error \\
\end{longtable}

\subsubsection{Dynamic subsystem or model}\label{dynamic-subsystem-or-model}

The dynamic subsystem consists of expressions of how the state variables or stocks, \(\boldsymbol{x}\) change over time, \(\boldsymbol{dx(t)/dt}\) or \(\boldsymbol{dx}\) for simplicity, along with the initial value of the stocks, \(\boldsymbol{x}_{t_0}\). The rate of change \(\boldsymbol{dx}\) is then defined as a column vector as a polynomial combination of coefficients, \(\beta_{1,i,j} \in B_1\) and exponents \(\gamma_{1,i,j} \in \Gamma_1\) of \(\boldsymbol{y}\):

\[
\boldsymbol{dx} = \begin{pmatrix}
dx_1 \\
\vdots \\
dx_m
\end{pmatrix} = 
\begin{pmatrix}
\beta_{1,1,1}y_1^{\gamma_{1,1,1}} + & \cdots & + \beta_{1,1,n}y_n^{\gamma_{1,1,n}}  \\
\vdots & \ddots & \vdots \\
\beta_{1,m,1}y_1^{\gamma_{1,m,1}} +& \cdots & + \beta_{1,m,n}y_1^{\gamma_{1,m,n}} 
\end{pmatrix}, 
\]

where \(\beta_{i,j}, \gamma_{i,j} \in \mathbb{R}\) are constants, \(y_i\) is a static variable that is a combination of \(\boldsymbol{x}\) and \(\boldsymbol{y}\), \(\boldsymbol{B_1}\) and \(\boldsymbol{\Gamma_1}\) are defined as follows:

\[
\boldsymbol{B_1} = 
  \begin{pmatrix}
  \beta_{1,1,1} & \cdots & \beta_{1,1,n} \\
  \vdots  & \ddots & \vdots \\
  \beta_{1,m,1} & \cdots & \beta_{1,m,n}
  \end{pmatrix} \; 
  \mathrm{and} \;
\boldsymbol{\Gamma_1} = 
  \begin{pmatrix}
  \gamma_{1,1} & \cdots & \gamma_{1,n} \\
  \vdots  & \ddots & \vdots \\
  \gamma_{m,1} & \cdots & \gamma_{m,n}
  \end{pmatrix}, 
\]

We can then define the function \(f\) that maps static variables to the rate of change as,

\[
\tag{1}
f:\boldsymbol{B_1} \times \boldsymbol{\Gamma_1} \times \boldsymbol{y} \rightarrow \boldsymbol{dx}
\].

The initial conditions are defined as the column vector,

\[
\tag{2}
\boldsymbol{x_{t_0}} = \begin{pmatrix}
x_{1_{t_0}} \\
\vdots \\
x_{m_{t_0}}
\end{pmatrix}, 
\]

where \(x_{i_{t_I}}\) is the initial value of the \(i\)-th state variable or stock at time \(t_I\). Note that while \(\boldsymbol{x_{t_I}}\) tends to be a vector of scalar values, it is not uncommon in system dynamics for \(\boldsymbol{x_{t_I}}\) to be defined by a set of equations that describe the initial values in terms of model parameters for under the condition when the initial rates of change of the system known, e.g., when a system is initially in a dynamic equilibrium.

\subsubsection{Static subsystem or model}\label{static-subsystem-or-model}

The static subsystem \(\boldsymbol{y}\) is the sum of three components, (a) the sum terms involving the state variables, \(\boldsymbol{x}\), (b) static components, \(\boldsymbol{y}\), and (c) two-way interactions, \(\boldsymbol{y} \otimes \boldsymbol{y}\).\footnote{The \(\otimes\) symbol is the outer product of two vectors \(\boldsymbol{x} = (x_1, x_2, \cdots, x_m)\) and \(\boldsymbol{y} = (y_1, y_2, \cdots, y_n)\) where \(\boldsymbol{y^T}\) is the transpose of \(\boldsymbol{y}\) and,
  \[
  \boldsymbol{x} \otimes \boldsymbol{y^T} = \begin{pmatrix}
  x_1y_1 & x_1y_2 & \cdots & x_1y_n \\
  \vdots & \vdots & & \vdots \\
  x_my_1 & x_my_2 & \cdots & x_my_n
  \end{pmatrix}.
  \]}

We define \(\beta_{2,i,j} \in B_2\), \(\beta_{3,i,j} \in B_3\), and \(\beta_{3,i,j} \in B_4\) as coefficients of \(\boldsymbol{x}\), \(\boldsymbol{y}\) and two-way interactions \(\boldsymbol{y} \otimes \boldsymbol{y}\), and \(\gamma_{2,i,j} \in \Gamma_2\), \(\gamma_{3,i,j} \in \Gamma_3\) to be exponents of the latent variables \(\boldsymbol{x}\) and \(\boldsymbol{y}\).

We define \(\beta_{2,i,j} \in B_2\), \(\beta_{3,i,j} \in B_3\), and \(\beta_{3,i,j} \in B_4\) as coefficients and \(\gamma_{2,i,j} \in \Gamma_2\) and \(\gamma_{3,i,j} \in \Gamma_3\) to be exponents of the latent variables \(\boldsymbol{x}\) and \(\boldsymbol{y}\).\footnote{We do not need to define exponents for the interaction terms because we can always define a static variable in terms of a power of another static variable}:

\[
\begin{split}
\boldsymbol{y} = \begin{pmatrix}
y_1 \\
\vdots \\
y_i \\
\vdots \\
y_n \\
\end{pmatrix}
 = 
& \begin{pmatrix}
  \beta_{2,1,1}x_1^{\gamma_{2,1,1}} + \cdots  + b_{2,1,m}x_m^{\gamma_{2,1,m}} \\
  \vdots \\
  \beta_{2,i,1}x_1^{\gamma_{2,i,1}} + \cdots  + \beta_{2,i,m}x_m^{\gamma_{2,i,m}} \\
  \vdots \\
  \beta_{2,n,1}x_1^{\gamma_{2,n,1}} + \cdots + \beta_{2,n,m}x_m^{\gamma_{2,n,m}} \\
  \end{pmatrix} +
\\
&  
  \begin{pmatrix}
  \beta_{3,1,1} y_1^{\gamma_{3,1,1}} + \cdots  + \beta_{3,1,n}y_n^{\gamma_{3,1,1}} \\
  \vdots \\
  \beta_{3,i,1}y_1^{\gamma_{3,i,1}} + \cdots  + \beta_{3,i,n}y_n^{\gamma_{3,i,n}} \\
  \vdots \\
  \beta_{3,n,1}y_1^{\gamma_{3,n,1}} + \cdots + \beta_{3,n,n}y_n^{\gamma_{3,n,n}} \\
  \end{pmatrix} + \\
&    
  \begin{pmatrix}
  \beta_{4,1,1,1}y_1y_1 +  \beta_{4,1,1,2}y_1y_2 + \cdots  + \beta_{4,1,n}y_1y_n + 
  \beta_{4,1,2,2}y_2y_2 + \cdots + \beta_{4,1,n,n}y_ny_n\\
  \vdots \\
  \beta_{4,i,1,1}y_1y_1 + \beta_{4,i,1,2}y_1y_2 + \cdots  + \beta_{4,i,1,n}y_1y_n + 
  \beta_{4,i,2,2}y_2y_2 + \cdots + \beta_{4,i,n,n}y_ny_n\\
  \vdots \\
  \beta_{4,n,1,1}y_1y_1 + \beta_{4,n,1,2}y_1y_2 + \cdots  + \beta_{4,n,n}y_1y_n + 
  \beta_{4,n,2,2}y_2y_2 + \cdots + \beta_{4,n,n,n}y_ny_n \\
  \end{pmatrix},
\end{split}
\]
where, for simplicity, we restrict \(\beta_{4,i,j,j}=0\) since the interaction term \(y_jy_j\) is already included in \(\beta_{3,j,j}\) with the exponent \(\gamma_{3,j,j}=2\).

If we then define \(\boldsymbol{B_2}\), \(\boldsymbol{B_3}\),and \(\boldsymbol{B_4}\) as
follows:

\[
\boldsymbol{B_2}  = 
  \begin{pmatrix}
    \beta_{2,1,1} & \cdots & \beta_{2,1,m} \\
    \vdots &  & \vdots \\
    \beta_{2,n,1} & \cdots & \beta_{2, n,m} \\
  \end{pmatrix}, 
\boldsymbol{B_3}  = 
    \begin{pmatrix}
    \beta_{3,1,1} & \cdots & \beta_{3,1,n} \\
    \vdots &  & \vdots \\
    \beta_{3,n,1} & \cdots & \beta_{3,n,n} \\
  \end{pmatrix},  \mathrm{and} \;
\boldsymbol{B_4}  = 
    \begin{pmatrix}
    \beta_{4,1,1,1} & \cdots & \beta_{4,1,n,n} \\
    \vdots &  & \vdots \\
    \beta_{4,n,1,1} & \cdots & \beta_{4,n,n,n} \\
  \end{pmatrix}.
\]

And,

\[
\boldsymbol{\Gamma_2}= 
  \begin{pmatrix}
  \gamma_{2,1,1} & \cdots & \gamma_{2,1,m} \\
  \vdots &  & \vdots \\
  \gamma_{2,n,1} & \cdots & \gamma_{2,n,m} \\
  \end{pmatrix}  \mathrm{and} \;
\boldsymbol{\Gamma_3}  = 
  \begin{pmatrix}
  \gamma_{3,1,1} & \cdots & \gamma_{3,1,n} \\
  \vdots &  & \vdots \\
  \gamma_{3,n,1} & \cdots & \gamma_{3,n,n} \\
  \end{pmatrix},
\]

where the function \(g\) maps latent variable vectors \(\boldsymbol{x}\) and \(\boldsymbol{y}\) to \(\boldsymbol{y}\):

\[
\tag{3}
g:\boldsymbol{B_2} \times \boldsymbol{B_3} \times \boldsymbol{B_4} \times \boldsymbol{\Gamma_2} \times \boldsymbol{\Gamma_3} \times \boldsymbol{x} \times \boldsymbol{y} \rightarrow \boldsymbol{y}.
\]

\begin{figure}

{\centering \includegraphics[width=0.2\linewidth]{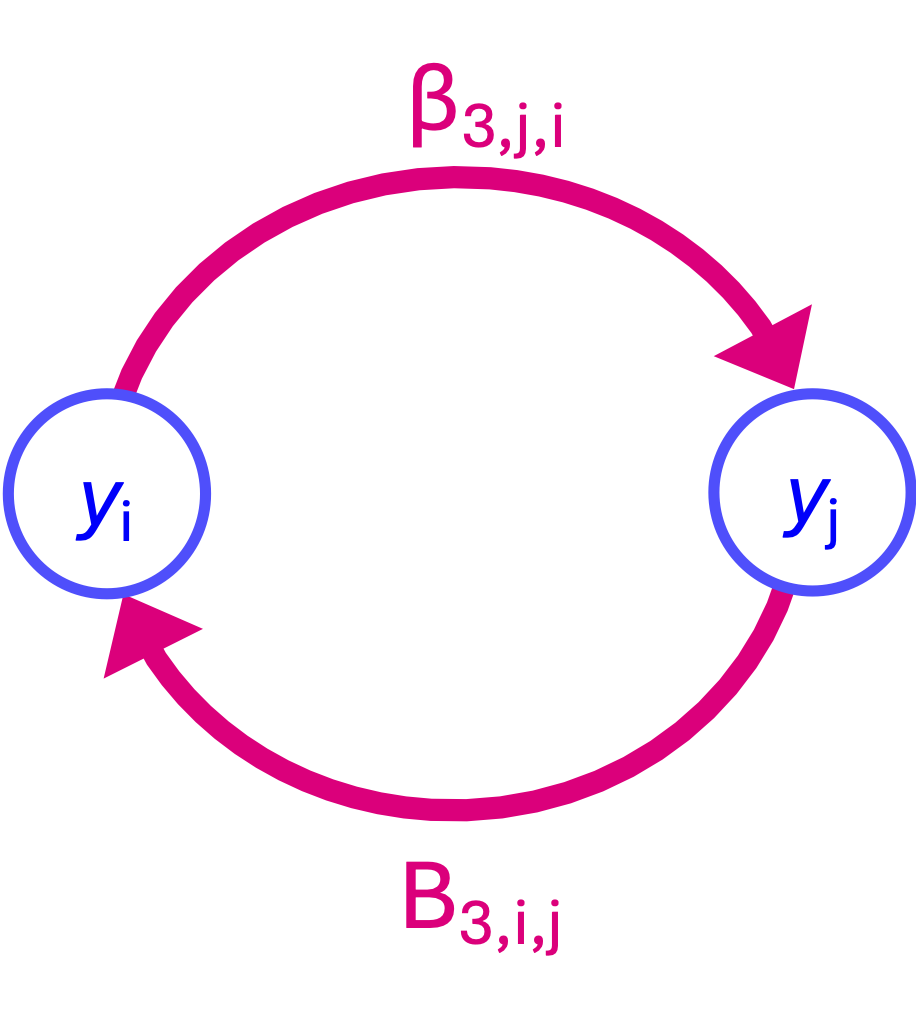} 

}

\caption{nonrecursive relationship between $y_i$ and $y_j$}\label{fig:diag-nonrecursive}
\end{figure}

There are several things to note about the static subsystem. First, it is more general than the conventional system dynamics model by allowing for non-recursive or simultaneous relationships between static variables, e.g., situations where there exist \(i,j \in [1,n]\) such that \(\beta_{3,i,j}\) and \(\beta_{3,j,i}\) are nonzero as shown in Figure \ref{fig:diag-nonrecursive}. In general, system dynamics models exclude nonrecursive or simultaneous equations as part of the static subsystem. There are exceptions, for example, as Ventana Vensim includes a built-in function for estimating simultaneous equations, however the general principle in system dynamics is that every feedback loop contain at least one state variable.

Second, for system dynamics models, we will generally require \(\boldsymbol{B_3}\) and \(\boldsymbol{B_4}\) to be of lower triangular matrices with a zero diagonals since one of the basic principles of system dynamics is that every feedback loop must have at least one stock variable. Hence, there can be no cycles between static variables \(\boldsymbol\), which implies that the equations defining relationships between static variables take the following lower triangular form:

\[
\boldsymbol{B_3} = 
  \begin{pmatrix}
  0 & \cdots  & & &  0 \\
  \beta_{3,2,1} & 0 & \cdots & & 0 \\
  \beta_{3,3,1} & \beta_{3,3,2} & 0 & \cdots & 0 \\
  \vdots &  & & & \vdots \\
  \beta_{3,n,1} &  \beta_{3,n,2} & \cdots & \beta_{3,n,n-1} & 0 \\
  \end{pmatrix}, \;  \mathrm{and} \;
\boldsymbol{B_4}  = 
  \begin{pmatrix}
  0 & \cdots  & & &  0 \\
  \beta_{4,2,1} & 0 & \cdots & & 0 \\
  \beta_{4,3,1} & \beta_{4,3,2} & 0 & \cdots & 0 \\
  \vdots &  & & & \vdots \\
  \beta_{4,n,1} & \beta_{4,n,2} & \cdots & \beta_{4,n,n-1} & 0 \\
  \end{pmatrix}.
\]

There is generally no such restriction in SEM, which allows for the specification of non-recursive or reciprocal relationships.

\subsubsection{Measurement subsystem or model}\label{measurement-subsystem-or-model}

The measurement subsystem maps the latent vectors \(\boldsymbol{x}\) and \(\boldsymbol{y}\) to the column vector \(\boldsymbol{z}\) of \(p\) indicators over \(q\) observations at \(t \in [t_1,\cdots,t_q]\) such that we have the following \(p\) by \(q\) data matrix of observations:

\[
\boldsymbol{Z} = 
  \begin{pmatrix}
  z_1(t_1) & \cdots & z_1(t_q) \\
  \vdots. & & \vdots \\
  z_p(t_1) & \cdots & z_p(t_q)
  \end{pmatrix}.
\]

To map our dynamic and static latent variables to our indicators, we allow for latent variables to be mapped to one or more indicators as well as having information delays between latent variables and observations across time. For simplicity, however, we assume that the measurement error \(\epsilon_i\) associated with each indicator \(z_i\) is constant for each indicator across time, \(\epsilon_i(t_j)=\epsilon_i(t_k)\) for all \(j,k \in [1,\cdots,q]\), which is a characteristic of good measurement. In future work, this assumption can be relaxed to consider the sensitivity of methods to measurement error that varies across time or correlated with other terms in a model.

To develop our measurement model, we start with \(t_1\), extend this to the \(t_2\) and \(t_3\) cases, and then our general form with delays. For \(t_1\) we then have:

\[
\begin{split}
z_1(t_1) & = \lambda_{x_1(t_1)z_1(t_1)}x_1(t_1) + \cdots +
             \lambda_{x_m(t_1)z_1(t_1)}x_m(t_1) + 
             \lambda_{y_1(t_1)z_1(t_1)}y_1(t_1) + \cdots
             \lambda_{y_n(t_1)z_1(t_1)}y_n(t_1) + \cdots \epsilon_{z_1} \\
          &   = \sum_{i=1}^m \lambda_{x_i(t_1)z_1(t_1)}x_i(t_1) + 
                \sum_{i=1}^n \lambda_{y_i(t_1)z_1(t_1)}y_i(t_1) + \epsilon_{z_1} \\
z_2(t_1) & =    \sum_{i=1}^m \lambda_{x_i(t_1)z_2(t_1)}x_i(t_1) + 
                \sum_{i=1}^n \lambda_{y_i(t_1)z_2(t_1)}y_i(t_1) + \epsilon_{z_2} \\    
                \vdots \\
z_p(t_1) & = \sum_{i=1}^m \lambda_{x_i(t_1)z_p(t_1)}x_i(t_1) + 
             \sum_{i=1}^n \lambda_{y_i(t_1)z_p(t_1)}y_i(t_1) + \epsilon_{z_p}. 
\end{split}
\]

For \(t_2\), the indicator variables can be influenced by both the present value of latent variables at \(t_2\) and the previous value at \(t_1\), hence:

\[
\begin{split}
z_p(t_2) = & \sum_{i=1}^m \lambda_{x_i(t_2)z_p(t_2)}x_i(t_2) + 
             \sum_{i=1}^m \lambda_{x_i(t_1)z_p(t_2)}x_i(t_1) + \\
          &  \sum_{i=1}^n \lambda_{y_i(t_2)z_p(t_2)}y_i(t_2) + 
             \sum_{i=1}^n \lambda_{y_i(t_1)z_p(t_2)}y_i(t_1) + \epsilon_{z_p}. 
\end{split}
\]

For \(t_3\), the indicator variables can be influenced by both the present value of latent variables at \(t_3\) and the previous values at \(t_2\) and \(t_1\), hence:

\[
\begin{split}
z_p(t_3) = & \sum_{i=1}^m \lambda_{x_i(t_3)z_p(t_3)}x_i(t_3) + 
             \sum_{i=1}^m \lambda_{x_i(t_2)z_p(t_3)}x_i(t_2) + 
             \sum_{i=1}^m \lambda_{x_i(t_1)z_p(t_3)}x_i(t_1) + \\
          &  \sum_{i=1}^n \lambda_{y_i(t_3)z_p(t_3)}y_i(t_3)
             \sum_{i=1}^n \lambda_{y_i(t_2)z_p(t_3)}y_i(t_2) + 
             \sum_{i=1}^n \lambda_{y_i(t_1)z_p(t_3)}y_i(t_1) + \epsilon_{z_p}. 
\end{split}
\]
We can generalize this for \(t_k\) as the double sums,

\[
\tag{4}
\begin{split}
z_p(t_k) = & \sum_{j=1}^k \sum_{i=1}^m \lambda_{x_i(t_j)z_p(t_k)}x_i(t_j) +
             \sum_{j=1}^k \sum_{i=1}^n \lambda_{y_i(t_j)z_p(t_k)}y_i(t_j)
            + \epsilon_{z_p}. 
\end{split}
\]

Equation 4 represents the relationship between the indicator \(z_p\) and latent variables \(x\) and \(y\) with lagged observations corresponding to the discrete observations from \(t_1\) to \(t_q\) consistent with an SEM measurement model organized around discrete observations. However, system dynamics models are continuous in time and the delays between observations and latent indicators are continuous, i.e., they are not restricted to incremental delays between discrete observations (and rarely are). To address this, we introduce the term \(\theta \in [0,+\infty]\) to represent the delay between the observation \(z_i\) and value of the latent variable where the expression \(x_j(t_k-\theta_{i,j})\) means the value of \(x_j\) at \(t_k\) delayed by \(\theta_{i,j}\) for \(x_j\). Similar to our assumption that measurement errors for an indicator are constant across time, we make the simplifying assumption that the delays \(\theta_i\) are also constant across time.

We thus arrive at our general expression for our indicator \(z_p(t_k)\) as,

\[
\tag{5}
\begin{split}
z_p(t_k) = & \sum_{j=1}^k \sum_{i=1}^m \lambda_{x_i(t_j)z_p(t_k)}x_i(t_j-\theta_{i,p}) +
             \sum_{j=1}^k \sum_{i=1}^n \lambda_{y_i(t_j)z_p(t_k)}y_i(t_j-\theta_{i,p})
            + \epsilon_{z_p}. 
\end{split}
\]

Bringing this all together into a more general form, we can define,

\[
\begin{split}
\boldsymbol{Z} & = 
  \begin{pmatrix}
  z_1(t_1) & \cdots & z_1(t_q) \\
  \vdots & & \vdots \\
  z_p(t_1) & \cdots & z_p(t_q)
  \end{pmatrix} \\
  \boldsymbol{\Lambda_x^i} & = 
  \begin{pmatrix}
  \lambda_{x_1(t_1)z_i(t_1)} & \cdots  & \lambda_{x_1(t_1)z_i(t_q)} \\
  \vdots & & \vdots \\
  \lambda_{x_m(t_1)z_i(t_1)} & \cdots  & \lambda_{x_m(t_q)z_i(t_q)}
  \end{pmatrix} \\
  \boldsymbol{\Lambda_y^i} & = 
  \begin{pmatrix}
  \lambda_{y_1(t_1)z_i(t_1)} & \cdots  & \lambda_{x_1(t_1)z_i(t_q)} \\
  \vdots & & \vdots\\
  \lambda_{y_n(t_1)z_i(t_1)} & \cdots  & \lambda_{y_n(t_q)z_i(t_q)}
  \end{pmatrix} \\
  \boldsymbol{\Theta_x} & = 
  \begin{pmatrix}
  \theta_{1,1} & \cdots & \theta_{1,p} \\
  \vdots & & \vdots \\
  \theta_{m,1} & \cdots & \theta_{m,p} 
  \end{pmatrix} \\
  \boldsymbol{\Theta_y} & = 
  \begin{pmatrix}
  \theta_{1,1} & \cdots & \theta_{1,p} \\
  \vdots & & \vdots \\
  \theta_{n,1} & \cdots & \theta_{n,p} 
  \end{pmatrix}.
\end{split}
\]
where the function \(h\) maps our latent variables \(\boldsymbol{x}\) and \(\boldsymbol{y}\) to our observations or data matrix, \(\boldsymbol{Z}\):

\[
\tag{6}
h:\boldsymbol{\Lambda_{x}^i \times \Lambda_{y}^i \times \Theta_{x} \times \Theta_{y} \times x \times y \rightarrow Z}.
\].

\subsubsection{General model}\label{general-model}

The general model can then be described as a function \(f\),\(g\), and \(h\) along with the initial conditions \(\mathbf{x_{t_0}}\) over the time horizon or interval \([t_I,t_f]\) as shown in Table 2.

\textbf{Table 2.} General model

\begin{longtable}[]{@{}
  >{\raggedright\arraybackslash}p{(\columnwidth - 2\tabcolsep) * \real{0.2542}}
  >{\raggedright\arraybackslash}p{(\columnwidth - 2\tabcolsep) * \real{0.7458}}@{}}
\toprule\noalign{}
\begin{minipage}[b]{\linewidth}\raggedright
Subsystem
\end{minipage} & \begin{minipage}[b]{\linewidth}\raggedright
Function
\end{minipage} \\
\midrule\noalign{}
\endhead
\bottomrule\noalign{}
\endlastfoot
Dynamic & \(f:\boldsymbol{B_1} \times \boldsymbol{\Gamma_1} \times \boldsymbol{y} \rightarrow \boldsymbol{dx}\) and \(\boldsymbol{x_{t_0}}\) \\
Static & \(g:\boldsymbol{B_2} \times \boldsymbol{B_3} \times \boldsymbol{B_4} \times \boldsymbol{\Gamma_2} \times \boldsymbol{\Gamma_3} \times \boldsymbol{x} \times \boldsymbol{y} \rightarrow \boldsymbol{y}\) \\
Measurement & \(h:\boldsymbol{\Lambda_{x}^i \times \Lambda_{y}^i \times \Theta_{x} \times \Theta_{y} \times x \times y \rightarrow Z}\) \\
\end{longtable}

As a function, \(f\), \(g\), and \(h\) can be implemented in any programming language and used to simulate a system dynamics model with \(\boldsymbol{x_{t_0}}\) initial conditions over \([t_I,t_f]\). Specifying a model is then a matter of defining the matrices in Table 2. Although appearing complex on the surface, in the next section we illustrate how models in system dynamics and structural equation modeling can be specified using this general model using a relatively small set of expressions.

\section{Examples}\label{examples}

In this section, we provide a few illustrative examples of how to specify various models using our proposed general model. The first illustrates the ``Limits to Growth'' system dynamics model and the second the ``Industrialization and Political Democracy'' structural equation model. The third is a system dynamics ``concept'' model of a hypothetical childhood vaccination program. Concept models are small models developed for the specific purpose of introducing system dynamics conventions and the structure behavior relationship in a content area (Richardson, 2013). The ``Childhood Vaccination'' concept model was developed and used in \emph{Salud Mesoamerica} 2015 Initiative (Munar et al., 2015). The last example is a system dynamics model of systems thinking and team performance that builds on a confirmatory factor analysis (CFA) and exploratory factor analysis (EFA) of the Systems Thinking Scale Revised (Davis and Stroink, 2016) using SEM.

\subsection{SD ``Limits to Growth'' model}\label{sd-limits-to-growth-model}

The ``Limits to Growth'' model (Figure 6) has one latent state variable representing the population, two latent flows, three constants, and three latent auxiliary variables as shown in the figure below. Translating this into a general model requires adding an additional static ``auxiliary'' variable \(y_9\) to facilitate interactions between the dynamic variable \(x_1\) and static variables. We also take advantage of the restriction in system dynamics that static variables cannot form feedback loops, which allows us to define constants by setting the exponent to zero and coefficient to the value of the constant, i.e., \(y_i=c\) implies \(\beta_{2,i,i}=c\) and \(\gamma_{2,i,i}=0\). Table 3 below provides the specification for specific terms. All other elements of the matrices in the general form are zero.

\begin{figure}
\centering
\includegraphics[width=0.5\textwidth,height=\textheight]{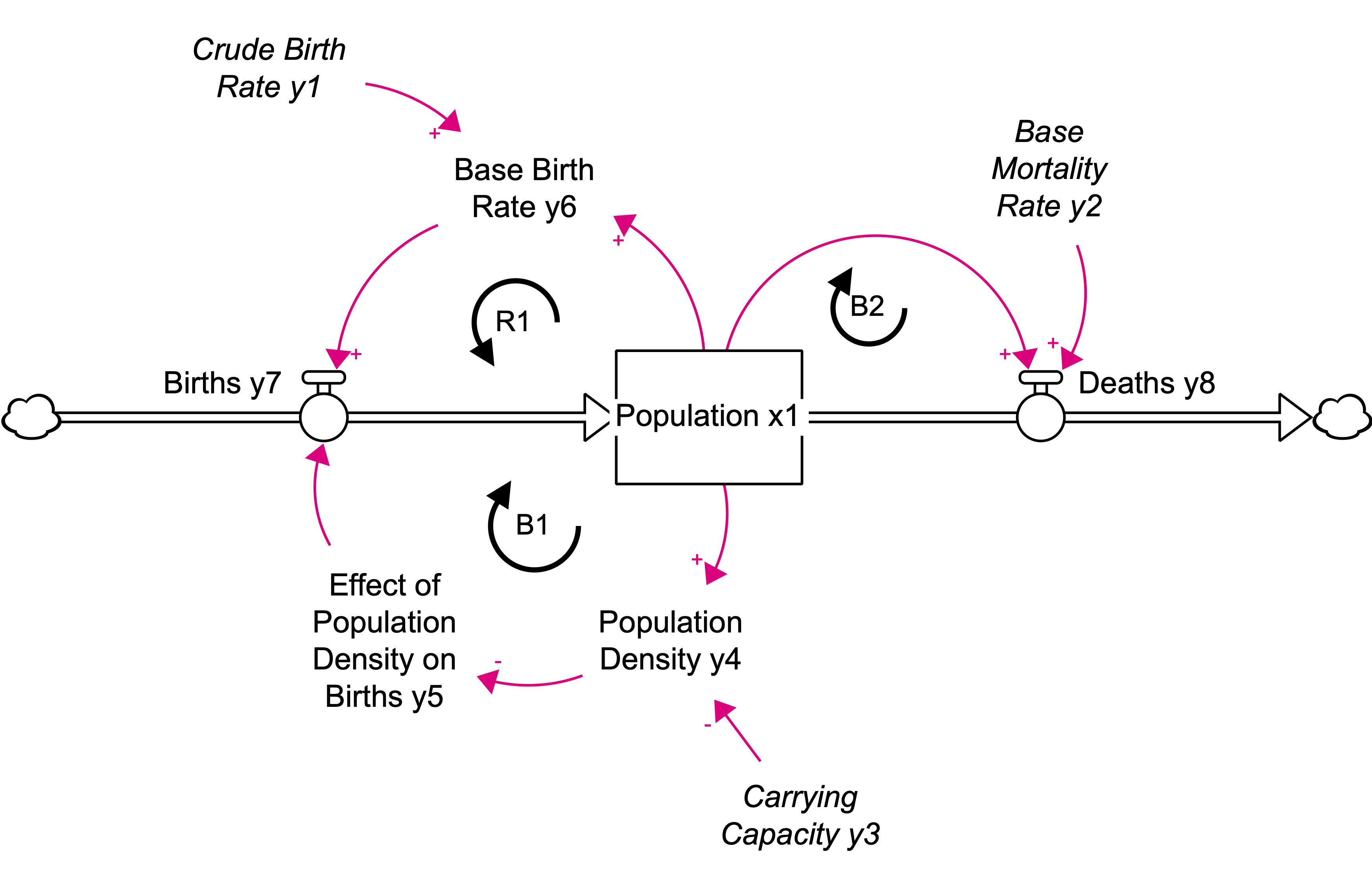}
\caption{``Limits to Growth'' model}
\end{figure}

\textbf{Table 3.} ``Limits to Growth'' variables, symbols and equations

\begin{longtable}[]{@{}
  >{\raggedright\arraybackslash}p{(\columnwidth - 6\tabcolsep) * \real{0.3056}}
  >{\centering\arraybackslash}p{(\columnwidth - 6\tabcolsep) * \real{0.2500}}
  >{\centering\arraybackslash}p{(\columnwidth - 6\tabcolsep) * \real{0.2222}}
  >{\centering\arraybackslash}p{(\columnwidth - 6\tabcolsep) * \real{0.2222}}@{}}
\toprule\noalign{}
\begin{minipage}[b]{\linewidth}\raggedright
Variable
\end{minipage} & \begin{minipage}[b]{\linewidth}\centering
Symbol
\end{minipage} & \begin{minipage}[b]{\linewidth}\centering
Equation
\end{minipage} & \begin{minipage}[b]{\linewidth}\centering
Model Specification
\end{minipage} \\
\midrule\noalign{}
\endhead
\bottomrule\noalign{}
\endlastfoot
Population(\(t_0\)) & \(x_{t_0}\) & 1.0 & \(x_{t_0}=1.0\) \\
Population & \(x_1\) & \(y_7 - y_8\) & \(\beta_{1,1,7}=1\), \(\beta_{1,1,8}=-1\) \\
Crude Birth Rate & \(y_1\) & 0.15 & \(\beta_{3,1,1}=0.15\), \(\gamma_{3,1,1}=0\) \\
Base Mortality Rate & \(y_2\) & 0.05 & \(\beta_{3,2,2}=0.05\), \(\gamma_{3,2,2}=0\) \\
Carrying Capacity & \(y_3\) & 100 & \(\beta_{3,3,3}=100\), \(\gamma_{3,3,3}=0\) \\
Population Density & \(y_4\) & \(x_1/y_3\) & \(\beta_{2,4,1}=1\), \(\gamma_{2,4,1}=1\), \(\beta_{3,4,3}=1\), \(\gamma_{3,4,3}=-1\) \\
Effect of Population Density on Births & \(y_5\) & \(1- y_4\) & \(\beta_{3,5,5}=1\), \(\gamma_{3,5,5}=0\), \(\beta_{3,5,4}=-1\), and \(\gamma_{3,5,4}=1\) \\
Base Birth Rate & \(y_6\) & \(x_1 \cdot y_1\) & \(\beta_{4,6,1,9} =1\) \\
Births & \(y_7\) & \(y_6 \cdot y_5\) & \(\beta_{4,7,5,6} =1\) \\
Deaths & \(y_8\) & \(x_1 \cdot y_2\) & \(\beta_{4,8,1,2} =1\) \\
(auxiliary) & \(y_9\) & \(x_1\) & \(\beta_{2,9,1}=1\), \(\gamma_{2,9,1}=1\) \\
\end{longtable}

\subsection{SEM ``Industrialization and Political Democracy'' model}\label{sem-industrialization-and-political-democracy-model}

We illustrate an SEM measurement model using the well-known example of structural equation modeling (Bollen, 1989), the Industrialization and Political Democracy dataset with 75 observations and 11 observed variables (Figure 7). It is worth noting that while this dataset represents two points in time (1960 and 1965), it is essentially estimated as a cross-sectional dataset.

\begin{figure}
\centering
\includegraphics[width=0.5\textwidth,height=\textheight]{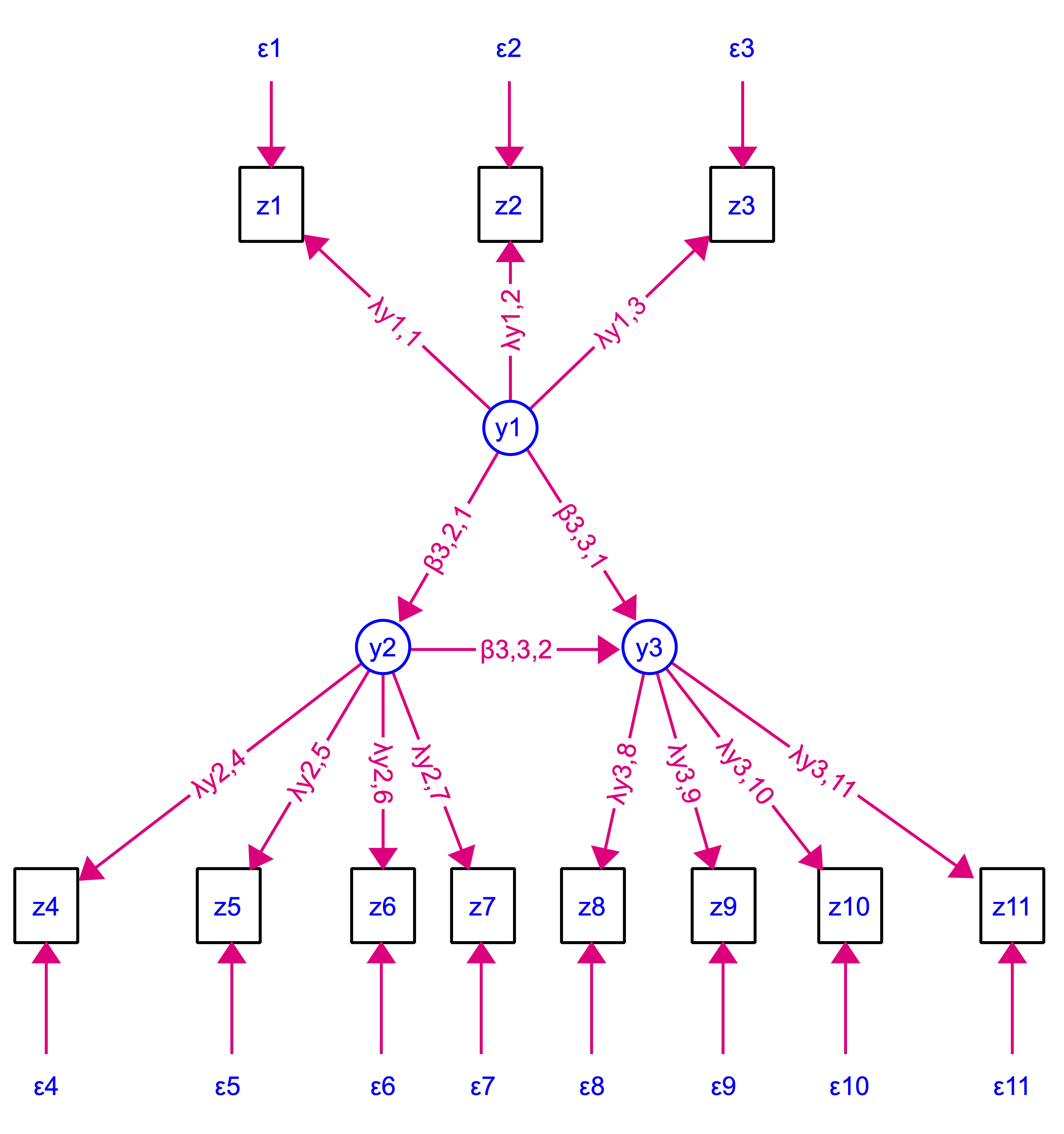}
\caption{``Industrialization and Political Democracy'' model}
\end{figure}

\textbf{Table 4.} Industrialization and Political Democracy

\begin{longtable}[]{@{}
  >{\raggedright\arraybackslash}p{(\columnwidth - 4\tabcolsep) * \real{0.3929}}
  >{\centering\arraybackslash}p{(\columnwidth - 4\tabcolsep) * \real{0.3214}}
  >{\centering\arraybackslash}p{(\columnwidth - 4\tabcolsep) * \real{0.2857}}@{}}
\toprule\noalign{}
\begin{minipage}[b]{\linewidth}\raggedright
Variable
\end{minipage} & \begin{minipage}[b]{\linewidth}\centering
Symbol
\end{minipage} & \begin{minipage}[b]{\linewidth}\centering
Model Specification
\end{minipage} \\
\midrule\noalign{}
\endhead
\bottomrule\noalign{}
\endlastfoot
ind60 & \(y_1\) & \\
dem60 & \(y_2\) & \(\beta_{3,2,1}\) \\
dem65 & \(y_3\) & \(\beta_{3,3,2} + \beta_{3,3,1}\) \\
GDP in 1960 & \(z_{1}\) & \(\lambda_{y_1z_1}\) \\
Per capita energy consumption in 1960 & \(z_{2}\) & \(\lambda_{y_1z_2}\) \\
Labor force in industry in 1960 & \(z_{3}\) & \(\lambda_{y_1z_3}\) \\
Freedom of the press in 1960 & \(z_{4}\) & \(\lambda_{y_2z_4}\) \\
Freedom of political opposition in 1960 & \(z_{5}\) & \(\lambda_{y_2z_5}\) \\
Fairness of elections in 1960 & \(z_{6}\) & \(\lambda_{y_2z_6}\) \\
Effectiveness of the elected legislature in 1960 & \(z_{7}\) & \(\lambda_{y_2z_7}\) \\
Freedom of the press in 1965 & \(z_{8}\) & \(\lambda_{y_3z_8}\) \\
Freedom of political opposition in 1965 & \(z_{9}\) & \(\lambda_{y_3z_9}\) \\
Fairness of elections in 1965 & \(z_{10}\) & \(\lambda_{y_3z_{10}}\) \\
Effectiveness of the elected legislature in 1965 & \(z_{11}\) & \(\lambda_{y_3z_{11}}\) \\
\end{longtable}

\subsection{SD ``Childhood Vaccinations'' concept model}\label{sd-childhood-vaccinations-concept-model}

In the next example, the model simulates a hypothetical childhoodl vaccination program from \(t_I=0\) to \(t_F=5\) years.

\includegraphics[width=0.85\textwidth,height=\textheight]{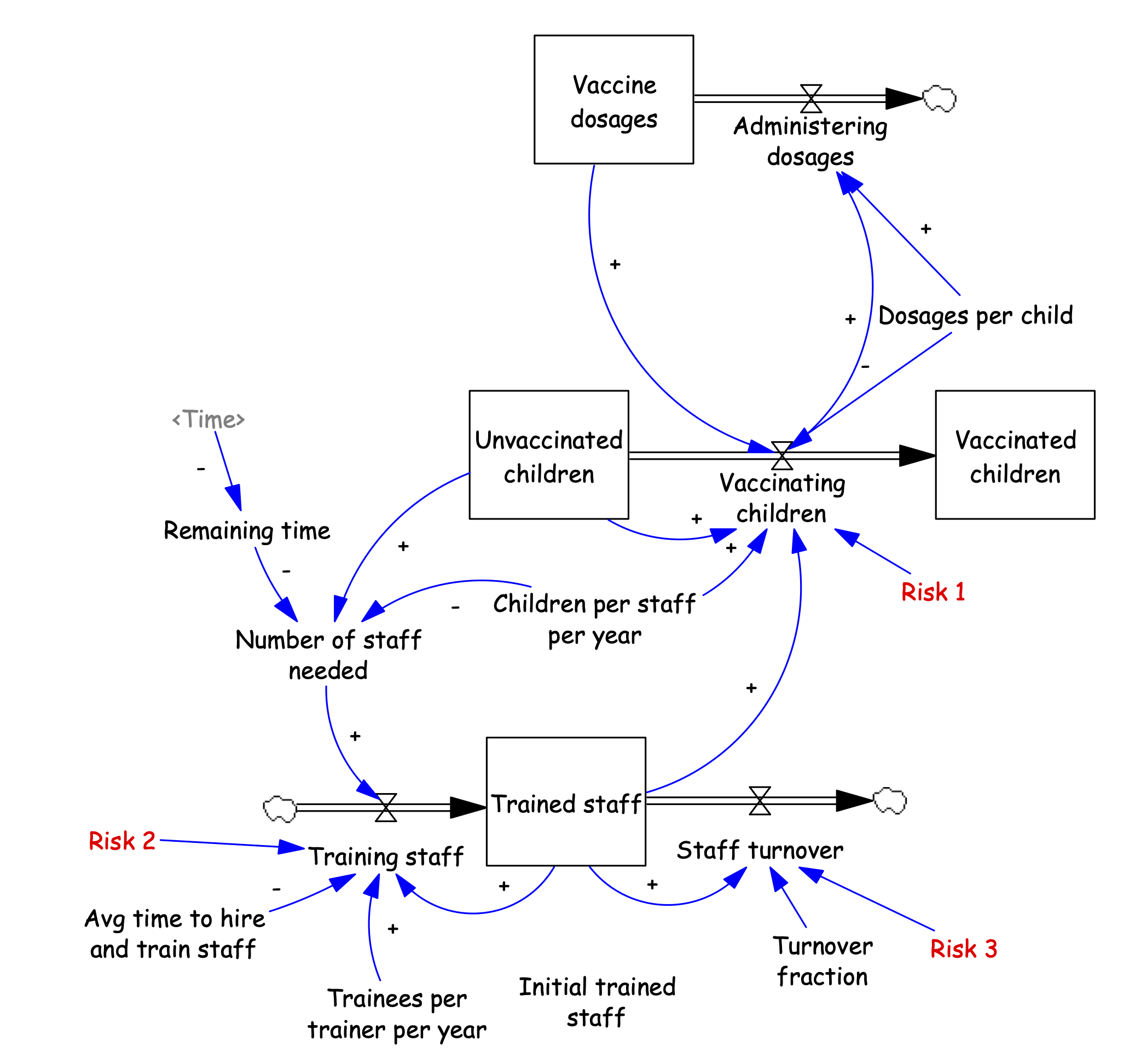}
\textbf{Table 5.} ``Childhood Vaccinations'' concept model

\subsection{SD-SEM ``Systems Thinking and Team Performance'' model}\label{sd-sem-systems-thinking-and-team-performance-model}

This last examples focuses on combining SD and SEM within the same study to better understand the relationship between systems thinking and team performance in an undergraduate course. Of particular interest here is the use of SEM to establish the latent structure of ``systems thinking'' using the \emph{Systems Thinking Scale Revised (STSR)} (Davis and Stroink, 2016). \emph{STSR} has been previously used and analyzed as a unidimensional measure of systems thinking (Davis and Stroink, 2016; Davis et al.~2018; Ballew et al 2019).

\begin{figure}
\centering
\includegraphics[width=1\textwidth,height=\textheight]{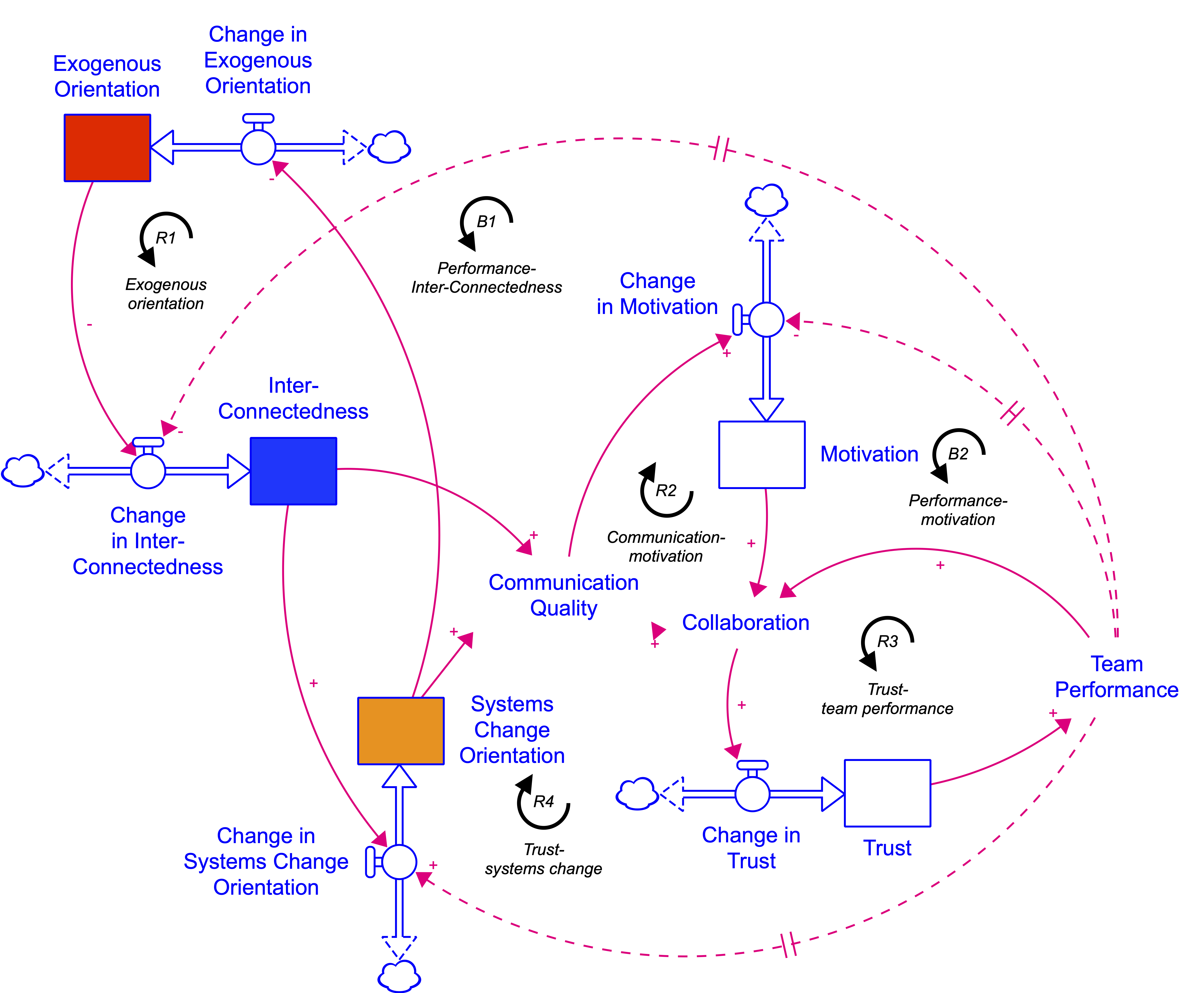}
\caption{``Systems Thinking and Team Performance'' model}
\end{figure}

\section{Discussion}\label{discussion}

In this paper, we have highlighted the differences between structural equation modeling and system dynamics, and then sought to develop a general framework in an effort to bridge the ``unavoidable a priori'' between two different approaches to causal modeling. The general model (Table 2) defines a set of functions that maps a model specification into a data frame or matrix of observed variables over time.

The generality of the approach means that various models from system dynamics and structural equation modeling can be mapped into the same general model. We illustrate this by showing how two standard models, one from system dynamics and another from structural equation modeling, can be represented in this general model. From this, one can represent a wide range of both system dynamics and structural equation models including hybrid models that bring the two approaches together in more interesting ways.

The main contribution is not, however, to suggest a parsimonious synthesis of the two approaches, but rather, to define a mathematical space that can be used to systematically generate and compare methods for causal modeling over well-defined distributions of systems. That a mathematical space is complicated shouldn't be surprising, but that it can be characterized at all within a relatively finite and small set of dimensions that we suspect can cover most of the usual cases in both methods is interesting. Complicated, yes, but not a bridge between two methods that is impossible to build.

Defining the mathematical space for bridging these two methods provides a means to generate systems in a systematic way and conduct simulation studies to explore the implications of various assumptions on causal inferences from modeling and data. This becomes even more critical today when we consider the implications of implementing AI/ML models in systems that impact society because, more than ever before, we need to understand how the assumptions we make and hence our biases might translate into a algorithmic biases. Presently, we have ideas about how we might be able to mitigate some of these biases and arrive at more accurate and ethical models using causal maps and system dynamics in particular, but without a solid understanding of how methods relate, we are at risk of blindly letting the unavoidable a priori be an excuse for the ``unintended'' consequences of harm caused by poorly designed AI/ML.

If one of the main contributions of system dynamics is to help us think better about the dynamically complex systems we are in and trying to manage, and our method is grounded in a deep mathematical understanding of our natural and social world as dynamic, it behooves us as a discipline to seek out and mathematically reconcile differences with other methods. We do this because there is an underlying principle in system dynamics about how we reconcile competing claims through mathematical models that capture the logic of our views on the world and data. We do this because we care about a fundamental idea of Truth.

\section{Acknowledgements}\label{acknowledgements}

This research was supported by the National Center For Advancing Translational Sciences of the National Institutes of Health under Award Number RC2TR004518. The content is solely the responsibility of the authors and does not necessarily represent the official views of the National Institutes of Health.

\section{References}\label{references}

Archontoulis, S. V., \& Miguez, F. E. (2015). Nonlinear regression models
and applications in agricultural research {[}Article{]}. \emph{Agronomy Journal,
107}(2), 786-798. \url{https://doi.org/10.2134/agronj2012.0506}

Ballew, M. T., Goldberg, M. H., Rosenthal, S. A., Gustafson, A., \& Leiserowitz, A. (2019, 2019/04/23). Systems thinking as a pathway to global warming beliefs and attitudes through an ecological worldview. \emph{Proceedings of the National Academy of Sciences, 116}(17), 8214-8219. \url{https://doi.org/10.1073/pnas.1819310116}

Berger, P. L., \& Luckmann, T. (1966). \emph{The social construction of
reality: a treatise in the sociology of knowledge.} Doubleday. Benjamin,
R. (2019). \emph{Race after technology: abolitionist tools for the new Jim
code.} Polity Press.

Boker, S. M., \& Wenger, M. J. (2007). \emph{Data analytic techniques for
dynamical systems.} Lawrence Erlbaum Associates.

Bollen, K. A. (1989). \emph{Structural equations with latent variables.} John
Wiley \& Sons, Inc.

Bunge, M. (1997). Mechanism and explanation. \emph{Philosophy of Social
Sciences, 27}, 410-465.

Cohen, J., Cohen, P., West, S. G., \& Aiken, L. S. (2003). \emph{Applied
multiple regression/correlation analysis for the behavior sciences (3rd
ed.)}. Lawrence Erlbaum Associates.

Davis, A. C., \& Stroink, M. L. (2016, 2016/07/01). The Relationship between Systems Thinking and the New Ecological Paradigm. \emph{Systems Research and Behavioral Science, 33}(4), 575-586.

Davis, A. C., Leppanen, W., Mularczyk, K. P., Bedard, T., \& Stroink, M. L. (2018, 2018/03/01). Systems Thinkers Express an Elevated Capacity for the Allocentric Components of Cognitive and Affective Empathy. \emph{Systems Research and Behavioral Science, 35}(2), 216-229. \url{https://doi.org/https://doi.org/10.1002/sres.2475}

Forrester, J. W. (1968). Industrial dynamics-after the first decade.
\emph{Management Science 14}(7): 398--415.

Forrester, J. W. (1980). Information sources for modeling the national
economy. \emph{Journal of the American Statistical Association, 75}(371),
555-566.

Forrester, J. W. (1990). \emph{Principle of systems.} Pegasus Communications,
Inc.~(1971)

Gunzler, D. D., Perzynksi, A. T., \& Carle, A. C. (2021). \emph{Structural
equation modeling for health and medicine.} Chapman \& Hall/CRC.

Hertz, H. (2003). \emph{The principles of mechanics presented in a new form.}
Dover Publications. Publisher description
\url{http://www.loc.gov/catdir/enhancements/fy0615/2003062517-d.html}

Hovmand, P. S. (2003). Analyzing dynamic systems: A comparison of
structural equation modeling and system dynamics modeling. In B. H.
Pugesek, A. Tomer, \& A. v. Eye (Eds.), \emph{Structural equation modeling:
Applications in ecological and evolutionary biology} (pp.~212-234).
Cambridge University Press.

Hovmand, P. S., \& Chalise, N. (2015). Simultaneous Linear Estimation
Using Structural Equation Modeling. In H. Rahmandad, R. Oliva, \& N. D.
Osgood (Eds.), \emph{Analytical Methods for Dynamic Modelers} (pp.~71-94).
The MIT Press.

Kuhlberg, J. A., Headen, I., Ballard, E. A., \& Martin Jr, D. (2023).
Advancing Community Engaged Approaches to Identifying Structural Drivers
of Racial Bias in Health Diagnostic Algorithms. arXiv preprint
arXiv:2305.13485.

Lakatos, I. (1970). Falsification and the methodoogy of scientific
research programmes. In I. Lakatos \& A. Musgrave (Eds.), \emph{Criticism and
the Growth of Knowledge} (pp.~91-196). Cambridge University Press.

Levine, L. R., Sell, M. V., \& Rubin, B. (1992). System dynamics and the
analysis of feedback processes in social and beahvioral systems. In L.
R. Levine \& H. E. Fitzgerald (Eds.), \emph{Analysis of Dynamic Psychological
Systems, Vol 1.: Basic Approaches to General Systems, Dynamic Systems,
and Cybernetics} (pp.~145-266). Plenum Press.

Martin, D. \& Kinney, D. (2024). Loop Polarity Analysis to Avoid
Underspecification in Deep Learning. \url{https://arxiv.org/abs/2309.10211}

Martin, D., \& Moore, A. (2020). AI Engineers Need To Think Beyond
Engineering. \emph{Harvard Business Review Blog} (October, 28, 2020).

Meadows, D. H. (1976). The unavoidable a priori. 1976 Conference of the
System Dynamics Society Geilo, Norway.

Meehl, P. (1990). Appraising and Amending Theories: The Strategy of
Lakatosian Defense and Two Principles That Warrant It. \emph{Psychological
Inquiry, 1}(2), 108 - 141.

Munar, W., Hovmand, P. S., Fleming, C., \& Darmstadt, G. L. (2015). Scaling-up impact in perinatology through systems science: Bridging the collaboration and translational divides in cross-disciplinary research and public policy {[}Review{]}. \emph{Seminars in Perinatology, 39}(5), 416-423. \url{https://doi.org/10.1053/j.semperi.2015.06.003}

Muthén, L. K., \& Muthén, B. O. (2012). \emph{Mplus. The comprehensive
modelling program for applied researchers: user's guide, 5}.

National Research Council. (2004). \emph{Measuring Racial Discrimination.}
The National Academies Press. \url{https://doi.org/doi:10.17226/10887}

Nganyu Tanyu, D., Ning, J., Freudenberg, T., Heilenkötter, N.,
Rademacher, A., Iben, U., \& Maass, P. (2023, 2023/08/25). Deep learning
methods for partial differential equations and related parameter
identification problems. \emph{Inverse Problems, 39}(10), 103001.
\url{https://doi.org/10.1088/1361-6420/ace9d4}

Palm, W. J. (1983). \emph{Modeling, analsysis and control of dynamic systems}. John Wiley \& Sons.

Pearl, J., \& Mackenzie, D. (2018). \emph{The book of why: the new science of
cause and effect.} Basic Books.

Pearl, J. (2009). \emph{Causality: models, reasoning, and inference (2 ed.).}
Cambridge University Press.

powell, j. a. (2008). Structural racism: building upon the insights of
John Calmore. \emph{North Carolina Law Review, 66}(3), 791-816.

Richardson, G. P., \& Pugh, A. L. (1986). \emph{Introduction to system
dynamics modeling with DYNAMO.} MIT Press.

Richardson, G. P. (1997). Problems in causal loop diagrams. \emph{System
Dynamics Review, 13}(3), 247-252.

Richardson, G. P. (2013). Concept models in group model building. \emph{System Dynamics Review, 29}, 42-55.

Schweppe, F. C. (1973). \emph{Uncertain dynamic systems.} Prentice Hall.

Spearman, C. (1904). ``General Intelligence,'' objectively determined and
measured. \emph{The American Journal of Psychology, 15}(2), 201-292.

Sterman, J. D. (2000). \emph{Business dynamics: Systems thinking and modeling
for a complex world.} Irwin McGraw-Hill.

Tarka, P. (2018, 2018/01/01). An overview of structural equation
modeling: its beginnings, historical development, usefulness and
controversies in the social sciences. \emph{Quality \& Quantity, 52}(1),
313-354.

Wright, S. (1921). Correlation and causation. \emph{Journal of agricultural
research, 20}(7), 557-585.

Zeigler, B. P. (1976). \emph{Theory of Modeling and Simulation.} John Wiley \&
Sons.

\end{document}